\pdfoutput=1

\documentclass[11pt]{article}

\usepackage[final]{acl}

\usepackage{times}
\usepackage{latexsym}
\usepackage{float}
\usepackage{placeins}
\usepackage{afterpage}
\usepackage[T1]{fontenc}

\usepackage[utf8]{inputenc}

\usepackage{microtype}

\usepackage{inconsolata}

\usepackage{graphicx}

\usepackage{mathtools}

\usepackage{subcaption}
\usepackage{graphicx}
\usepackage{amsfonts} 
\usepackage{booktabs}
\usepackage{wrapfig}
\usepackage{tabularx}
\usepackage{array,multirow}

\usepackage[inline]{enumitem}

\let\originalparagraph\paragraph
\renewcommand{\paragraph}[2][.]{\originalparagraph{#2#1}}

\newcommand{\relation}[1]{\textsc{#1}}

\begin{document}

\newcommand{\model}[1]{\textsc{#1}}
\newcommand{\dataset}[1]{\textit{#1}}





\title{From Data to Knowledge: Evaluating How Efficiently Language Models Learn Facts}





\author{
    Daniel Christoph\footnotemark[2]
    \quad
    Max Ploner \footnotemark[2]\footnotemark[3]
    \quad
    Patrick Haller \footnotemark[2]
    \quad
    Alan Akbik \footnotemark[2]\footnotemark[3]
    \vspace*{2.5mm}\\
    \footnotemark[2] Humboldt-Universität zu Berlin\\
    \footnotemark[3] Science Of Intelligence \\
    \texttt{<firstname>.<lastname>@hu-berlin.de} \\ 
}

\maketitle

\begin{abstract}
Sample efficiency is a crucial property of language models with practical implications for training efficiency. In real-world text, information follows a long-tailed distribution. Yet, we expect models to learn and recall frequent and infrequent facts. Sample-efficient models are better equipped to handle this challenge of learning and retaining rare information without requiring excessive exposure. This study analyzes multiple models of varying architectures and sizes, all trained on the same pre-training data. By annotating relational facts with their frequencies in the training corpus, we examine how model performance varies with fact frequency. Our findings show that most models perform similarly on high-frequency facts but differ notably on low-frequency facts. This analysis provides new insights into the relationship between model architecture, size, and factual learning efficiency.
\end{abstract}

\section{Introduction}
With the continued advancement of language models (LMs), comparing different architectures across various tasks and evaluating their performance using appropriate metrics becomes increasingly essential. These comparisons offer valuable insights into each architecture's general strengths and limitations. Sample efficiency is a key property of LMs, as sample-efficient models require less training and are thus more cost-effective \citep{micheli2023transformerssampleefficientworldmodels}.
As the LM processes large text corpora during pre-training, we are interested in assessing how efficiently each model learns specific relational facts comprising a subject, relation, and object.

\begin{figure}[t]
\centering
\includegraphics[trim=6mm 5mm 6mm 5mm, width=\linewidth]{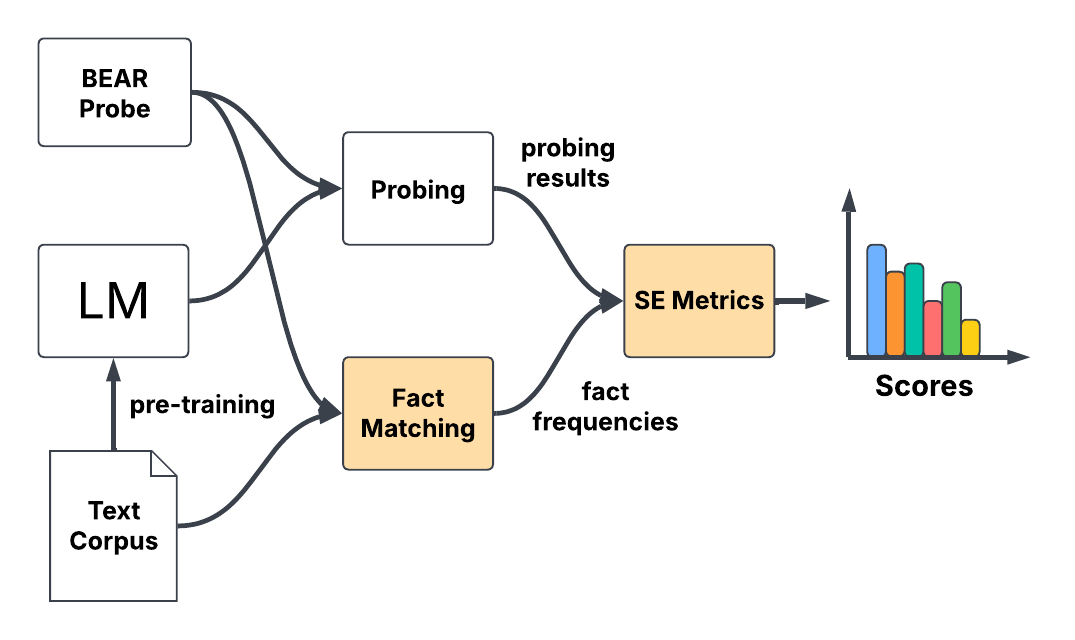}
\caption{Sample efficiency evaluation of LMs.}
\label{thesis-main-diagramm}
\end{figure}


A core question in this context is how different architectures handle the challenge of learning and retaining rare versus frequent facts. If two models are trained on the same dataset, their sample efficiency can be assessed by determining how often a fact must appear before each model successfully learns it \citep{botvinickReinforcementLearningFast2019, liu2023sampleefficientnlpmodelsrobust}.
Models that rely predominantly on frequent facts while struggling with rarer ones---an issue caused by the long-tailed distribution of information in natural text \citep{zhang2024systematicreviewlongtailedlearning}---are considered sample-inefficient.
Conversely, sample-efficient models should achieve higher accuracy on rare facts while maintaining strong performance on more common ones. To assess a model's factual knowledge, we use the BEAR probe \citep{wiland-etal-2024-bear}, which evaluates the model's ability to recall factual information across a wide range of subject-relation-object triples.

An LM's factual knowledge can be probed by passing statements into the model (e.g., ``\textit{The capital of Germany is ...}'') and evaluating its output to determine the represented knowledge of an LM \citep{robertsHowMuchKnowledge2020,kaloKAMELKnowledgeAnalysis2022,kandpalLargeLanguageModels2023}. BEAR enables evaluation of both causal and masked LMs by constructing multiple answer choices, where each instance is transformed into a set of natural language statements: One for each answer option (e.g., \textit{``Berlin''}, \textit{``Paris''}, \textit{``Buenos Aires''}, etc. for the relation \relation{has-capital} and the subject \textit{``Germany''}). The LM assigns log-likelihood scores to these statements, which are then ranked to determine the predicted answer.


Since BEAR contains no information about the pre-training data, it alone cannot be used to assess the sample efficiency.
To address this, we need to not only determine whether the LM can correctly recall a given fact but also how many times it encountered it during pre-training (in the following, we call these ``frequencies'').
To create a correct sample efficiency evaluation procedure, we require an approach to estimate frequencies of facts from BEAR within a text corpus used for pre-training (see Figure \ref{thesis-main-diagramm}). For this study, we employ a simple matching-based heuristic (see Section~\ref{sec:fact-matching}). Though unable to capture every occurrence of a fact, we assume it to be sufficiently accurate to predict the relative frequencies.

Given the information about how often an LM has encountered specific facts and whether it can recall them correctly, we must determine how to translate these fact-level data to a sample efficiency measure. Rather than estimating the point at which an LM transitions from not knowing to having learned the fact, we propose a more nuanced perspective: Measuring the incremental gain in factual knowledge as a function of the number of training samples. To operationalize this, we introduce two complementary metrics, which we use to quantify and compare the sample efficiency of different models over varying levels of fact exposure.

\paragraph{Contributions}
Our contributions can be summarized as follows. We

\begin{enumerate}
  \item Develop a framework to measure fact frequencies in text corpora efficiently and release counts for matched fact frequencies for a pre-training corpus,\footnote{The repository containing the fact frequencies and code can be found here: \href{https://github.com/Jabbawukis/sample-efficiency-evaluation}{github.com/Jabbawukis/sample-efficiency-evaluation}.}
  
  \item Propose a novel method for estimating sample efficiency using a model's prediction on factual questions given the number of supporting frequencies in the pre-training corpus and

  \item Compare models of three different architectures and varying sizes regarding their sample efficiency.

\end{enumerate}
\vspace{.2ex}

\section{Related Work}

\begin{figure}[t!]
    \includegraphics[width=\linewidth]{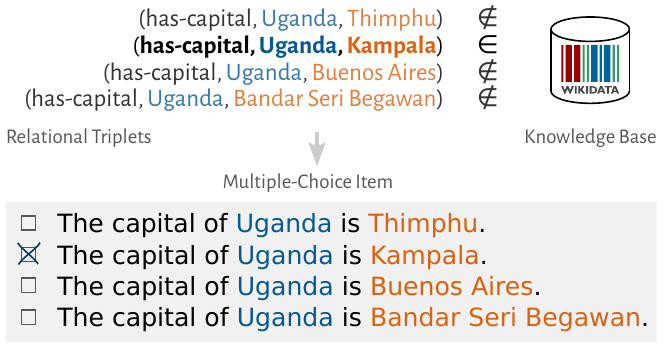}
    \caption{    
    In BEAR, one statement per answer option is passed to the LM (here using the template: ``The capital of [X] is [Y].'' and the subject ``Uganda''). The assigned sentence-level likelihoods are then used to rank the answer options \citep[figure from][]{plonerLMPUBQUIZComprehensiveFramework2024}.\label{fig:bear_eval_illustration}}
\end{figure}

\paragraph{Knowledge Probing}

\citet{petroni-etal-2019-language} introduced the influential \textit{LAMA} probe, which evaluates language models by generating sentences that express factual relations, masking the object entity, and prompting the model to fill in the blank. This method, however, only supports single-subword token predictions and is not compatible with non-masked models like GPT. Variants adapted for causal (autoregressive) language models exist \citep{robertsHowMuchKnowledge2020,kaloKAMELKnowledgeAnalysis2022,kandpalLargeLanguageModels2023}, but these cannot be used with masked LMs. To bridge this gap, BEAR \citep{wiland-etal-2024-bear} reformulates relation instances into multiple-choice items, creating natural language statements for each candidate answer, and probing the model to assign log-likelihoods to each of the statements. By comparing the statements with the highest likelihood with the true answer enables evaluation across both model types (see Figure~\ref{fig:bear_eval_illustration}). 
    
\paragraph{Sample Efficiency}
In the current literature, sample efficiency can be defined as the property of a model to achieve similar performance to comparable models on tasks while requiring less training data or achieving better results while training on the same data \citep{liu2023sampleefficientnlpmodelsrobust, lin2024largelanguagemodelsmake}. Reducing training time or data requirements is especially important when extensive data collection is expensive or impractical, which is especially challenging in domains with naturally low sample efficiency, potentially limiting real-world applicability~\citep{ijcai2018p820, feng2024sampleefficienthumanevaluationlarge}. 

\paragraph{Neural Scaling Laws}\label{sec:scaling-laws}

\citet{kaplanScalingLawsNeural2020} show that the test data's loss value depends on the pre-training data scale. Given that the model is sufficiently large and enough compute is available, it follows a power-law relationship, i.e. in a log-log plot the function appears roughly as a linear line with negative slope and can hence be modeled by a function of the form $y=x^{-k}$.

Subsequent studies extend these findings to transfer learning \citep{hernandezScalingLawsTransfer2021}, rigorously test this hypothesis, provide practical guidelines for optimal model-to-pre-training dataset size ratios \citep{hoffmannTrainingComputeOptimalLarge2022}, and propose methods for computing scaling laws using intermediate checkpoints \citep{choshenHitchhikersGuideScaling2024}.
Finally, \citet{godeyScalingLawsGeographical2024} identify power-law relationships related to encoded geographic knowledge and
\citet{luScalingLawsFact2024}, the most relevant to our study, examines model size and training time in fact memorization.

To our knowledge, no prior work has examined the direct relationship between fact frequencies in the pre-training data and the model’s ability to recall these facts.

\begin{figure*}[t]
\centering
\includegraphics[width=\linewidth]{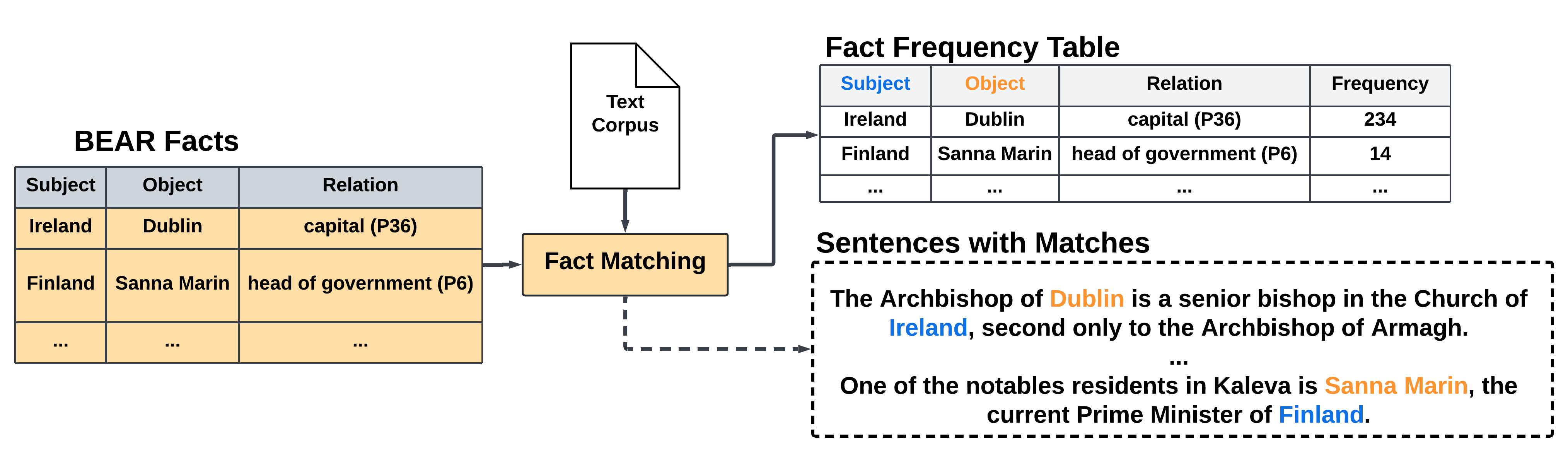}
\caption{Example fact frequency table constructed from a text corpus. A fact is counted if the subject and the object occur within a sentence, even if the sentence does not explicitly express the relation.}
\label{fig_process_dataset}
\end{figure*}

\section{Approach}
\label{approach}
To evaluate a model's sample efficiency, we employ a three-step approach. We build on BEAR and extend the probe by collecting fact frequencies (see Section~\ref{sec:fact-matching}) for a given pre-training corpus. We then train several LMs on this corpus (Section~\ref{sec:lm-training}). This way, we can estimate how often a model has encountered a specific fact during its pre-training (and at which point). 
In Section~\ref{sec:sm-evaluation}, we introduce two novel sample efficiency metrics which produce aggregated scores based on the model's response to each sample and the sample's frequency.

\subsection{Corpus Fact Frequency Statistics}
\label{sec:fact-matching}

To estimate how often a certain fact appears in the pre-training data, we look at single sentences and detect wether the fact is likely to be expressed within the sentence.
For simplicity, we only check if two entities (belonging to a specific fact triple) occur within the same sentence from the corpus. If so, we assume the relational fact is represented within the sentence \citep{mintz-etal-2009-distant}.

For example, given the sentence ``\textit{The \textnormal{Boeing 747} is a long-range wide-body airliner designed and manufactured by \textnormal{Boeing Commercial Airplanes} in the United States [...]}'', the occurrence of both entities ``\textit{Boeing 747}'' and ``\textit{Boeing Commercial Airplanes}'' can be observed and the two entities are assumed to be linked by the \relation{manufacturer} relation.
The entity ``\textit{Boeing Commercial Airplanes}'' in this example may also be referred to as simply ``\textit{Boeing}'' or ``\textit{Boeing commercial airplanes}''. Hence, it is crucial to account for potential aliases of entities and to discard case sensitivity. Once two relation entities have been identified within a sentence, the sentence is counted as a fact occurrence (see Figure~\ref{fig_process_dataset}).

We use rule-based lemmatization (for English language) and sentence-splitting (\textit{Sentencizer}) functionality provided by the spaCy Python library \cite{spacy2}. Lemmatization greatly improves the matching with the entity aliases. The approach is implemented in the \texttt{FactMatcherSimple} class in the repostory linked in the contributions.

\begin{figure}[t]
\includegraphics[width=\linewidth, trim=2mm 0 2mm 0]{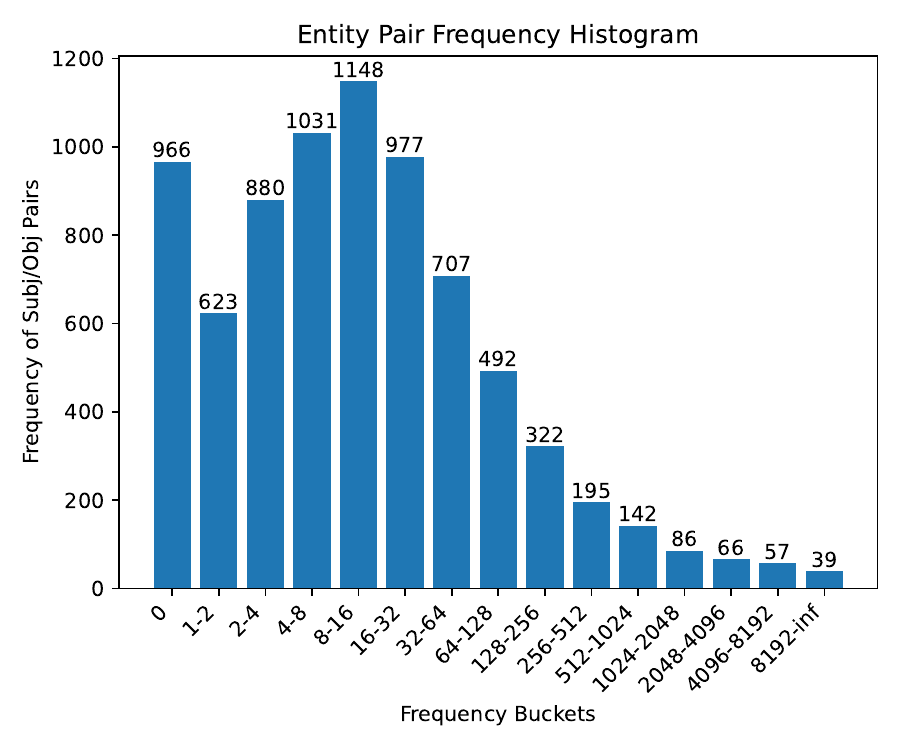}
\caption{Number of matches for BEAR facts in the English Wikipedia dump  \citep[\texttt{20231101.en}; ][]{wikidump}.}
\label{fig_approach_matchin_results}
\end{figure}

Selecting an appropriate corpus is crucial for generating useful fact-frequency statistics, as the chosen corpus must contain sufficient facts shared with the BEAR probe. If the text corpus lacks key information, entities from the BEAR probe may not appear with adequate frequency. To address this challenge, datasets derived from English Wikipedia articles, such as the Wikipedia dump language modeling dataset, can be utilized \citep{wikidump}. We applied this heuristic to the said corpus, and for better visualization, we placed each fact into a bucket relating to the overall frequency. The result is depicted in Figure~\ref{fig_approach_matchin_results}.

\subsection{Pre-Training the LMs}
\label{sec:lm-training}

We pre-train several language model (LM) architectures and sizes, targeting comparable language modeling quality (see Section~\ref{model_architecturs})
on approximately five billion tokens of Wikipedia text \citep[\texttt{20231101.en}; ][]{wikidump}.
For each model architecture, we train a small and a medium-sized model.
To enable fine-grained fact tracking and to closely monitor each model’s ability to recall facts over time, intermediate model checkpoints are saved and evaluated throughout training, allowing us to capture the learning dynamics in detail (see Section~\ref{slice_eval_explain}).

\subsection{Evaluating the Sample Efficiency}
\label{sec:sm-evaluation}

To measure sample efficiency, a common approach is to track the number of encounters a model has with a specific fact during training and continuously probe the model to record when it has answered the question relating to the fact correctly \citep{liu2023sampleefficientnlpmodelsrobust, lin2024largelanguagemodelsmake}.
However, since facts are usually not learned in isolation, e.g., facts not associated with a specific question may still contain enough information to enable the model to acquire the knowledge required to answer the question correctly or make educated guesses, this approach may not suffice. Additionally, the model may provide the correct answer at a specific moment in training but may later give the incorrect answer after it has processed more data, leading to a different outcome. There may not be a clear definition of learning a fact in a binary sense, as required.

To address these issues, we generalize this notion of sample efficiency: Instead of determining the critical point of knowledge acquisition, we conceptualize sample efficiency as the performance of correctly recalling facts as a function over the number of times the model has encountered this fact in the pre-training.

\subsubsection{Weighted Accuracy Score on Frequency Buckets}
\label{WASB}

A straightforward way is to measure the accuracy achieved on the facts of each frequency bucket (as illustrated in Figure~\ref{fig_approach_matchin_results}). This provides a good initial impression of an LM's performance on rare and frequent facts. However, the array of scores makes it difficult to compare multiple models or track an LM's sample efficiency throughout the training. Hence, we propose an additional metric to condense these results to a single score, substantially simplifying the comparison. A computationally simple approach takes a weighted average over the buckets, weighting buckets with lower frequencies higher to focus on rarer facts.
We propose the following weighting function based on the bucket $i$'s lower bound $l_i$:

\begin{equation*}
  w_i=\begin{cases}
    \exp(-\lambda l_i), & \text{if $l_i\ge1$}.\\
    0, & \text{otherwise}.
  \end{cases}
\end{equation*}

where $\lambda$ is set to $0.05$. The weight decreases with higher $l_i$, yielding $w_i\in \left[0,1\right)$, resulting in a declining impact of the high-frequency facts on the overall weighted accuracy (see Appendix Figure \ref{fig:weight-impact-bear}). The weighted accuracy is then calculated (with the accuracy score $\text{acc}_i$ on bucket $i$) as:

\begin{equation*}
    \frac{1}{\sum_{i=1}^{N}w_i}\sum_{i=1}^{N}w_i \cdot \text{acc}_i
\end{equation*}

If the fact has a particular frequency of $x$, we assign the fact to the bucket with a lower bound of $l_i$ and an upper bound of $u_i$  iff. $x \in \left[l_i,u_i\right)$ .

\subsubsection{Modeling the Probability of an LM to Answer Correctly}
\label{PSF_EXT_3}

A second approach is to apply a probabilistic interpretation and to treat sample efficiency as a key property of the function mapping the number of fact frequencies to the probability of the model recalling the fact accurately.
Within this framework, the threshold of the step function would represent the conventional notion of sample efficiency: The exact number of frequencies needed to give the correct answer consistently.

The step function may be ill-suited to model the actual probability of the model giving the correct answer. Instead, we propose to use a continuous function, where a higher slope of the function indicates a higher likelihood of the model learning a function and, thus, a higher sample efficiency. This approach eliminates the need to identify when a model has learned a specific fact by generalizing the evaluation to groups of facts rather than individual instances, potentially allowing for a more robust assessment of sample efficiency across varying levels of exposure in the training data.

We statistically model the probability of an LM correctly answering a question, given the number of frequencies of the related fact in the training data using a power scaling function \citep[see also the segment on neural scaling laws in Section~\ref{sec:scaling-laws};][]{kaplanScalingLawsNeural2020}: 

\begin{equation*}
    F(x) = 1 - \left(L_0 + \frac{x_0}{(1+x)^{\alpha_m}}\right)
\end{equation*}

Here, $x$ is the frequency of a fact, and $L_0$, $x_0$, and $\alpha$ are found by statistical fitting. While $L_0$ and $x_0$ are dataset dependent, there is one $\alpha_m$ per LM.

$\alpha_m$ controls the slope of the probability function: Higher values increase the probability per additional occurrence, indicating higher sample efficiency.

$L_0$ can be interpreted as the constant rate of error that is unavoidable, given the possibility that the BEAR probe contains errors (zero would indicate that the potential errors in the probe's question catalog do not influence the function).

$x_0$ is at least influenced by the fact-matching algorithm described in Section~\ref{sec:fact-matching}. Underestimating fact frequencies could result in a lower estimated $x_0$ value. Values lower than one indicate the LM's initial probability of correctly answering a fact can be $\ge0$, and values close to zero suggest an unexpectedly high probability, even though the fact frequency is zero.
Such a value might be produced due to the simplicity of the fact-matching heuristic or the learning of facts through other facts that hold helpful information for the fact in question or, in other words, educated guesses.

Representing LM $m$'s prediction on fact $i$ as $T_{m,i}$ (one if the model answered correctly, zero otherwise) yields a likelihood $p_{m,i}$ that the model makes the given prediction (given the modeled probability):
\begin{equation*}
p_{m,i} = T_{m,i} F(x_i) + (1 - T_{i,m})\left(1 - F(x_i)\right)
\end{equation*}

The overall probability of the predictions occurring as they have given the parameters $L_0$, $x_0$, and $\alpha_m$ is then given by:
\begin{equation*}
    P\left(L_0, x_0, \boldsymbol{\alpha} \right) = \prod_{m} \prod_{i} p_{m,i}
\end{equation*}

We maximize the joint probability (by minimizing the negative log-likelihood) over all BEAR probe facts and models. This yields the maximum likelihood estimate for our dataset-specific parameters $L_0$, $x_0$, and model-specific $\alpha_m$. LMs with a higher $\alpha_m$ value can be considered more sample efficient as they exhibit a higher increase in the probability of answering a factual item per observed sample.

\section{Empirical Evaluation}
Leveraging the proposed approach allows us to address the following questions: \begin{enumerate*}[label=(\arabic*)]

\item which model architecture demonstrates higher levels of sample efficiency,  
\item and how well a model recalls facts throughout the training.

\end{enumerate*}

\paragraph{LM Architecture Selection}
\label{model_architecturs}
Newer RNN-based architectures indicate advantages over transformer-based architectures in data-scarce scenarios and thus may indicate a higher sample efficiency~\citep{hallerBabyHGRNExploringRNNs2024}.
As the model architectures evaluated in this work consist of transformer-based \model{GPT2}~\citep{radford2019language} and \model{LLaMA}~\citep{touvron2023llamaopenefficientfoundation}, RNN-based \model{xLSTM} \citep{beck2024xlstmextendedlongshortterm} and state-space-based \model{Mamba2} \citep{mamba2}, the selected model architectures are well-suited for this study and may contribute to a deeper understanding of sample efficiency, particularly in the context of RNNs versus transformers, as well as broader trends across different architectural paradigms.

We train two groups of models. A small group with sizes around 200 million parameters, and a medium-sized group with around 400 million parameters. Due to limited resources, we are restricted to a limited set of training runs and LM sizes.

For model pre-training of the different model architectures, we use the models and trainer implemented in the Hugging Face \textit{transformers} library
\citep{wolfTransformersStateoftheArtNatural2020}.

\subsection{Sample Efficiency of Different LM Architectures}
\label{sec:exp_1}
Our first experiment compares the LMs' sample efficiency. Specifically, we evaluate the model's accuracy scores on each frequency bucket, apply the proposed metrics, and calculate the overall accuracy on all BEAR questions for comparison.

\subsubsection{Experimental Setup} Each model is trained on the same information-rich text corpus \citep{wikidump} using the same vocabulary (GPT2 tokenizer) and training parameters to ensure maximum comparability (see Appendix Table \ref{tab:training_params}). Each pre-training run took two to three days and was done on a single NVIDIA A100 (80GB) GPU. The models were evaluated using the proposed sample efficiency metrics (see Section~\ref{sec:sm-evaluation}). Additionally, each model was evaluated using several tasks from the language model evaluation harness \citep{evalharness}, including \dataset{winogrande}, \dataset{wsc273}, \dataset{lambada\_standard} and \dataset{pile\_10k} to test the model's general language modeling capabilities (see Appendix Table \ref{tab:lm_eval_harness_results}). 

\subsubsection{Results}
\label{results}

Table~\ref{tab:results_bear} reports the overall accuracy on all questions (ACC), the weighted accuracy score on the frequency buckets (WASB, see Section \ref{WASB}), and the optimized $\alpha_m$ values (see Section~\ref{PSF_EXT_3}) for the LMs in consideration (final state). The $L_0$ and $x_0$ values are optimized to 0.00 and 0.88, respectively. This indicates a base probability of a question being answered correctly by the model greater than zero and the general correctness of the BEAR probe question catalog.\footnote{
For BEAR-big, the resulting values for $L_0$ and $x_0$ are 0.0 and 0.92, respectively. The respective table (\ref{tab:results_bear_big}) can be found in Appendix~\ref{sec:additional_results}.} Going forward, we propose using the values we determined since $x_0$ and $L_0$ are dataset characteristics and not model-dependent (though future refinements using a larger set of models are possible).

\begin{table}[t]
    \centering\small
\begin{tabularx}{\linewidth}{cX|cccc}
 \toprule
 & Model & \#params & ACC & WASB & $\alpha_m$\\
 \midrule
 \parbox[t]{2mm}{\multirow{4}{*}{\rotatebox[origin=c]{90}{\textsc{small}}}}
 &\model{GPT2}   & 209M & 28.0\% & 21.8\% & 0.084\\
 &\model{LLaMA}  & 208M & \textbf{31.0\%} & \textbf{24.1\%} & \textbf{0.103}\\
 &\model{xLSTM}  & 247M & 28.1\% & 21.7\% & 0.086\\
 &\model{Mamba2} & 172M & \underline{28.6\%} & \underline{22.9\%} & \underline{0.087}\\
 \midrule
 \parbox[t]{2mm}{\multirow{4}{*}{\rotatebox[origin=c]{90}{\textsc{medium}}}}
 &\model{GPT2}   & 355M & 30.4\% & 24.0\% & 0.098\\
 &\model{LLaMA}  & 360M & \textbf{34.4\%} & \textbf{27.9\%} & \textbf{0.120}\\
 &\model{xLSTM}  & 406M & 30.7\% & 24.2\% & 0.100\\
 &\model{Mamba2} & 432M & \underline{32.1\%} & \underline{26.2}\% & \underline{0.106}\\
 \bottomrule
\end{tabularx}
    \caption{Resulting measures for LM's after pre-training on the complete corpus.}
    \label{tab:results_bear}
\end{table}

These results highlight two key observations. First, sample efficiency improves with increasing model size. Second, both \model{LLaMA} models consistently outperform other architectures with similar parameters.

\paragraph{Accuracies on Frequency Buckets}
Figure~\ref{fig:final_bucket_acc_all_models} in the appendix reports the model's accuracies on each frequency bucket. As Section~\ref{WASB} mentions, these scores provide an initial impression of the model's overall sample efficiency. Larger models achieve a higher accuracy score on the low to mid-frequency buckets ($\leq 128$). This finding indicates that larger LMs may learn less frequent facts better.

\paragraph{Accuracies on High Occurring Facts}
To verify this hypothesis, we split the facts into \textit{high-frequency} ($x \geq 1000$) and \textit{low-frequency} ($x < 1000$) facts and measure the accuracy on each of the splits.
Looking at these accuracies (in Table~ \ref{tab:acc_high_low_facts_results_bear}), we again observe an explicit ordering of the model performances in correlation with their size (as observed in Table \ref{tab:results_bear}) for low-frequency facts. However, the performance on high-frequency facts does not follow this trend.

Accuracies on high-occurring facts show less deviation between the models, as some small models achieve accuracy scores comparable to the medium models (e.g., small \model{GPT2} and medium \model{Mamba2}).
These findings show that larger LMs may not memorize high-frequency, possibly redundant facts significantly better than smaller models, in line with observations made by \citet{luScalingLawsFact2024}. 

\begin{table}[t]
    \centering\small
    \begin{tabularx}{\linewidth}{cX|ccc}
     \toprule
     &Model & \#params  &  $<$ 1024 & $\geq$ 1024\\
     \midrule
     \parbox[t]{2mm}{\multirow{4}{*}{\rotatebox[origin=c]{90}{\textsc{small}}}}
     &\model{GPT2}   & 209M & 26.2\% & \underline{83.4\%}\\
     &\model{LLaMA}  & 208M & \textbf{29.1\%} & \textbf{88.7\%}\\
     &\model{xLSTM}  & 247M & 26.4\% & 79.4\%\\
     &\model{Mamba2} & 172M & \underline{26.8\%} & 82.2\%\\
     \midrule
     \parbox[t]{2mm}{\multirow{4}{*}{\rotatebox[origin=c]{90}{\textsc{medium}}}}
     &\model{GPT2}   & 355M & 28.6\% & \textbf{87.5\%}\\
     &\model{LLaMA}  & 360M & \textbf{32.7\%} & \underline{85.4\%}\\
     &\model{xLSTM}  & 406M  & 29.0\% & 82.2\%\\
     &\model{Mamba2} & 432M & \underline{30.5\%} & 81.4\%\\
     \bottomrule
    \end{tabularx}
        \caption{Accuracy on high and low frequency facts on BEAR.}
        \label{tab:acc_high_low_facts_results_bear}
\end{table}

The results indicate that eliminating high-frequency facts or adjusting their influence on the final accuracy score to mitigate their impact may be necessary to measure sample efficiency effectively. This, however, may heavily depend on the dataset used for pre-training and may not always be required.
In some cases, the accuracy alone may suffice to distinguish sample-efficient from sample-inefficient models (also see Figure \ref{fig:accuracy_on_slices_bear_small} and \ref{fig:accuracy_on_slices_bear_big} in the appendix).

\subsection{Learning Dynamics}
\label{sec:exp_2}

\begin{figure}[t!]
    \begin{subfigure}{\linewidth}
        \includegraphics[width=\linewidth]{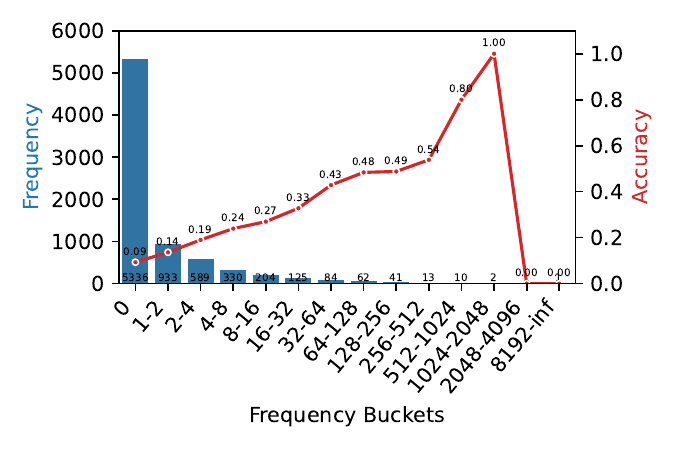}
        \caption{After training for 3,650 update steps}
        \label{fig:accuracy_slice_3650}
    \end{subfigure}
    
    \begin{subfigure}{\linewidth}
        \includegraphics[width=\linewidth]{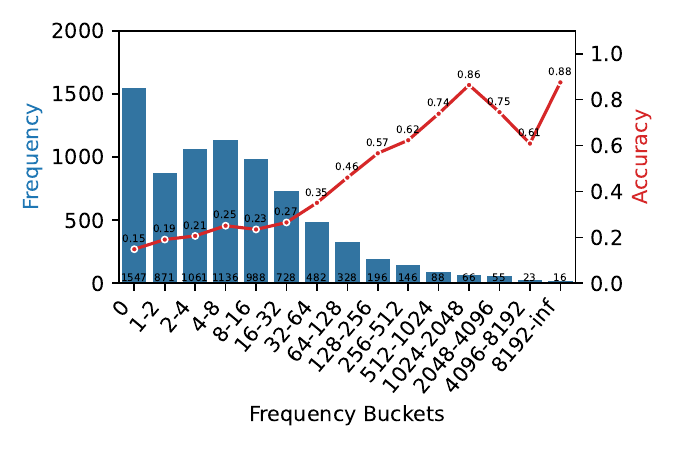}
        \caption{After training for 76,650 update steps}
        \label{fig:accuracy_slice_76650}
    \end{subfigure}
    
    \begin{subfigure}{\linewidth}
        \includegraphics[width=\linewidth]{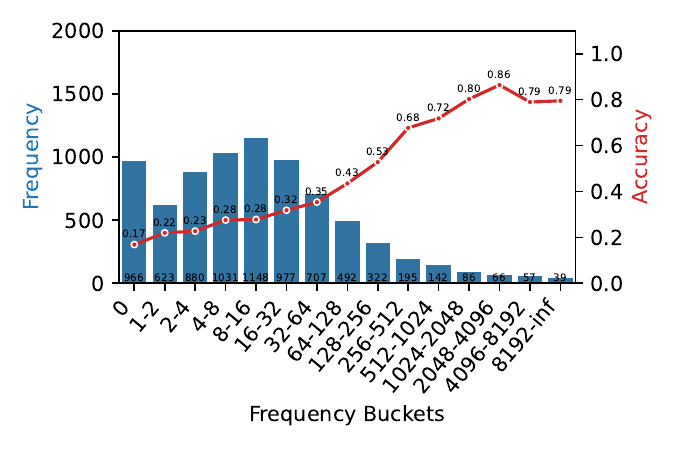}
        \caption{After training for 153,300 update steps}
        \label{fig:accuracy_slice_153300}
    \end{subfigure}
\caption{Accuracy on frequency buckets during training of Mamba2 with 432 million parameters. The top, middle, and bottom graphs depict the model's accuracy at the training's beginning, middle, and end.}
\label{fig:eval_on_slices}
\end{figure}

To investigate how the models acquire knowledge throughout the training, we probe the model periodically throughout the training. This also enables us to check if the proposed metrics are predictive of the final results: When do the bucket accuracies stabilize, and can we predict the final accuracy by extrapolating from a given checkpoint (knowing how often the facts will be seen in the data yet to be used during training)?

\subsubsection{Experimental Setup}
\label{slice_eval_explain}

The dataset is shuffled to account for a possible unbalanced distribution of data point sizes and was divided into 42 slices with 3650 steps per slice, with a train batch size of 32, gradient accumulation set to 8, and 934,840 rows per slice after tokenization on average ($934,840 \approx 8 \times 32 \times 3650$). Each slice is then processed using the fact-matching heuristic. We calculate the average\footnote{Using the mean instead of the slice-dependent number is not entirely accurate. However, since the variation between the slices (regarding training steps) is minimal, this simplification should not change the results.} number of training steps performed for each slice and save the model's state after a slice has been processed. Each state is then individually probed and evaluated based on the number of facts with specific frequencies the model has seen up until then. Probing each checkpoint for a single training run (i.e., 42 different model states) using BEAR-big (which includes BEAR as a subset) took approximately one day (single NVIDIA A100 (80GB) GPU). To substantially cut down the probing time, we recommend probing only using BEAR (without BEAR-big) and fewer checkpoints in practical settings.

\subsubsection{Results}
During training, we observe a gradual convergence toward specific accuracy scores for the lower frequency buckets relatively early, with increasingly smaller changes in the later stages of training. This indicates that a model's ability to learn a fact improves with the general learning of the meaning of language but remains relatively stagnant concerning frequency. This behavior is depicted in Figure~\ref{fig:eval_on_slices} (accuracy scores on frequency buckets during training of \model{Mamba2} with 432 million parameters and probed with BEAR).

Looking at the weighted accuracy scores (see Section \ref{WASB}) and $\alpha$-values (see Section \ref{PSF_EXT_3}) of the LMs over each slice, we observe a similar trend, with each model reaching a specific score early in training, with relatively minimal changes in the later stages of training (see Appendix Figure \ref{fig:weighted_accuracy_on_slices_bear} and \ref{fig:psf-ext3_optimized_alphas_bear}). However, the degree of increase in the scores during training seems to  depend on the model's overall capability to learn facts, as models with a higher final $\alpha$-value and weighted accuracy score show steeper increases, only reaching a stagnation point later in training.

\subsubsection{Correlation Between The Metrics}
\begin{figure}[t]
    \centering
    \begin{subfigure}{0.39\linewidth}
        \includegraphics[page=1, height=3.2cm]{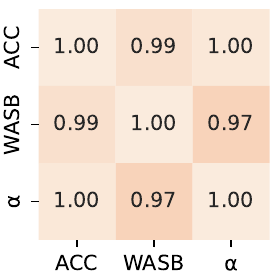}
        \caption{Final scores}
        \label{fig:corr_matrix_final_state_bear_small}
    \end{subfigure}
    \hfill 
    \begin{subfigure}{0.59\linewidth}
        \raggedleft
        \includegraphics[page=2,  height=3.2cm]{img/correlation/corr_both_bear_small_two_pages.pdf}
        \caption{Over all slices}
        \label{fig:corr_matrix_all_slices_bear_small}
    \end{subfigure}
    \caption{Correlation matrix of the final scores and over all slices.}
    \label{fig:corr_matrix_final_state_and_slices_bear_small}
\end{figure}

The proposed metrics indicate a clear trend: Larger models tend to outperform smaller models and are thus more sample-efficient, with exceptions observed in the \model{LLaMA} models, where the smaller model demonstrates competitive or superior performance compared to larger RNN-based models. This highlights the role of architectural efficiency beyond just scale. Additionally, the progression of the scores of each state of the models follows a similar trajectory in both proposed metrics, with minor variations in magnitude and fluctuations at specific points (see Appendix Figures ~\ref{fig:weighted_accuracy_on_slices_bear} and~\ref{fig:psf-ext3_optimized_alphas_bear}). This similarity suggests that both metrics are valid model performance indicators and can be used interchangeably or individually to assess sample efficiency. This results in a high correlation\footnote{Correlations were computed between metric scores across models at final training (Fig.~\ref{fig:corr_matrix_final_state_bear_small}; raw scores in Table~\ref{tab:results_bear}) using vectors \(v_M \in \mathbb{R}^{m \times 1}\). Correlations across all 42 training slices (Fig.~\ref{fig:corr_matrix_all_slices_bear_small}) use flattened vectors \(v_M \in \mathbb{R}^{m \times 42}\). Columns are sorted by correlation with overall accuracy.} between the proposed metrics across slices, while the correlation with the general accuracy is lower in comparison (see Figure~\ref{fig:corr_matrix_all_slices_bear_small}). On the other hand, we observe strong positive correlations for each metric for the final state (see Figure~\ref{fig:corr_matrix_final_state_bear_small}), as each metric sorts the model's final measurement similarly (larger models outperform smaller ones).

\subsection{Metric Robustness}
To further investigate the metrics' robustness to changes in the testing dataset's composition, we create two splits with 1000 facts from BEAR, each with a different frequency profile. Using these two splits, we aim to determine the impact of the different frequency profiles on the final metric.

Ideally, any testing dataset (no matter the makeup) could be used to estimate a model's sample efficiency based on the response patterns and information about the fact frequencies. We hypothesize that the fact frequencies highly impact the raw accuracy over the facts. In contrast, the weighted accuracy (WASB) and the modeling-based sample efficiency metric $\alpha$ might be less influenced by the sampling of the splits.

It should be noted that this assumes that the samples across the datasets are (on average) equally hard: The probability of the model to correctly predict the fact \emph{only} depends on the pre-training data and the model's sample efficiency (and not other difficulty factors).

\subsubsection{Experimental Setup}

For the \emph{low-frequency} split, we sample 80\% of the facts from facts with less than eight occurrences and the other 20\% from facts with eight occurrences or more. We do the opposite in the \emph{high-frequency} split (i.e., 80\% from facts with eight occurrences or more). The threshold must be set sufficiently to guarantee a strong bias within the split towards facts with a desired frequency. Otherwise, the split would be too close to the original data set. This can be achieved by calculating the median bucket lower bound for the fact counts, functioning as said threshold. We evaluate the final checkpoints of each model on these two new datasets and compute the different metrics.

\subsubsection{Results}

\begin{figure}[t]
    \centering

    \begin{subfigure}{\linewidth}
        \includegraphics[width=\linewidth]{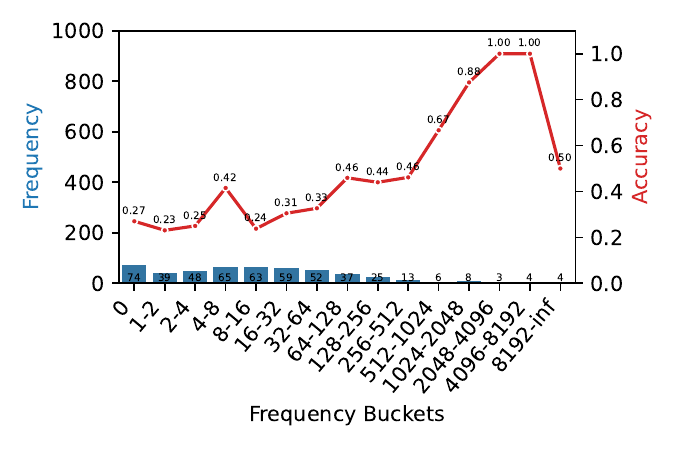}
        \caption{\emph{Low frequency} split}
        \label{fig:buckets-for-low-frequency-split}
    \end{subfigure}
    
    \begin{subfigure}{\linewidth}
        \includegraphics[width=\linewidth]{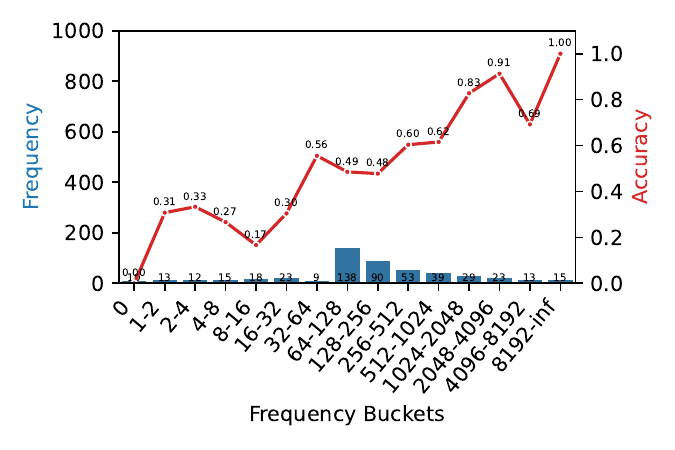}
        \caption{\emph{High frequency} split}
        \label{fig:buckets-for-high-frequency-split}
    \end{subfigure}
    \caption{Accuracy on frequency buckets after training of \model{Mamba2} for the two splits.}
    \label{fig:buckets-for-frequency-splits}
\end{figure}

The results are depicted in Figure~\ref{fig:all-metrics-for-frequency-splits} in the appendix.
The exemplary resulting frequency histogram and the accuracy for each bucket for \model{Mamba2} are shown in Figure~\ref{fig:buckets-for-frequency-splits}.

\paragraph{Accuracy}
The variation in the general accuracy among the models in the frequency splits is substantial. Compared to the scores on the complete dataset, the accuracy is lower if primarily low-frequency samples are selected and considerably higher in the high-frequency split (see Appendix Figure~\ref{fig:all-metrics-for-frequency-splits}). 

\paragraph{Weighted Accuracy (WASB)}


For the weighted accuracy measure on the frequency buckets for each model, the variation between the low and high frequency splits remains lower than the general accuracy. However, the weighted accuracy approach is limited by the need to adjust the buckets' resolution as more facts produce more robust results. Further investigation is needed to determine if there are robust ways to set the boundaries of the buckets based on the fact frequencies and the weights of each bucket based on these boundaries. This may lead to more robust measures where every subsample of the dataset can be used to estimate the overall performance. Additionally, calculating the weighted accuracy using accuracies on frequency buckets may result in less reliable scores when the number of samples within a bucket is too low. To address this, incorporating a confidence coefficient can help adjust for the increased uncertainty associated with smaller sample sizes.

\paragraph{$\alpha$-Sample Efficiency}
The $\alpha$-values exhibit the lowest variation between the low and high frequency splits (see Appendix Figure~\ref{fig:all-metrics-for-frequency-splits}). Thus, this modeling-based metric provides the highest robustness against fact frequency changes, resulting in the most reliable measures.

\section{Conclusion}
We presented a sample efficiency evaluation framework that compares LMs' ability to learn facts given a text corpus and the BEAR probe. The framework consists of a fact-matching algorithm that extracts fact frequency statistics from a sizable data set and two sample efficiency metrics. We trained several state-of-the-art LMs in a controlled setting, ensuring the validity of the evaluation, and provided a detailed analysis of the different architecture results.

The performance on high-frequency facts indicates less divergence between models regarding size. In contrast, performance on low-frequency facts demonstrates the increased sample efficiency gained with model size. The proposed metrics are capable of identifying the superiority in sample-efficiency of the transformer-based \model{LLAMA} models, achieving the highest scores in all metrics, with the state-space-based \model{Mamba2} models closing behind. 

The proposed metrics correlate strongly in respect to the final model stages as well as across the training.
This indicates that a different property is measured than in raw accuracy.
Additional experiments show, that the metrics are relatively robust to varying fact frequency distributions in pre-training data.
We believe the plausibility of the design choices together with these findings make the metrics strong candidates for measuring sample efficiency.

\section*{Limitations}
This work is limited to a simple fact-matching heuristic, as discussed in Section \ref{sec:fact-matching}. This heuristic produces sufficiently accurate statistics and provides a high degree of flexibility; however, more advanced heuristics, e.g., adding natural language processing pipelines such as entity linking, could produce more accurate fact occurrence counts, as they potentially reduce the possible mismappings of entities due to likely ambiguity or relation misidentification. Furthermore, the proposed probability function lower bound depends on $L_0$, validated empirically in this work (see Section \ref{results}). However, this initial $L_0$ value can change depending on the correctness of the probe (or the training text corpus), as significant errors and noise can alter the outcome of the measurements. Thus, further research could be conducted on the robustness of the metric in those scenarios. Finally, this work is limited to evaluating models of small to medium size. Whether the observed trend of increasing sample efficiency with model size holds for larger models exceeding one billion parameters remains open.

\makeatletter
\ifacl@finalcopy


\section*{Acknowledgments}

We thank all reviewers for their valuable comments. 
Alan Akbik and Max Ploner are supported under Germany’s Excellence Strategy ``Science of Intelligence'' (EXC 2002/1, project number 390523135).
Further, Alan Akbik and Patrick Haller are supported by the Deutsche Forschungsgemeinschaft (DFG, German Research Foundation) under Emmy Noether grant ``Eidetic Representations of Natural Language'' (project number 448414230).
\fi
\makeatother

\bibliography{references}

\clearpage
\appendix

\section{Pre-Training \& Model Configuration}
\label{sec:training_setup}

\begin{table}[ht]
    \centering\small
    \begin{tabularx}{\linewidth}{X|c}
        \toprule
        Parameter & Value \\
        \midrule
        per\_device\_train\_batch\_size & 32 \\
        gradient\_accumulation\_steps & 8 \\
        num\_train\_epochs & 1 \\
        weight\_decay & 0.1 \\
        warmup\_steps & 1000 \\
        lr\_scheduler\_type & cosine \\
        learning\_rate & 5e-4 \\
        fp16 & True \\
        \bottomrule
    \end{tabularx}
    \vspace*{-2mm}
    \caption{Training Hyperparameters.}
    \label{tab:training_params}
    \vspace*{-2mm}
\end{table}
\begin{table}[ht!]
    \centering\small
    \begin{tabularx}{\linewidth}{cX|c|c}
        \toprule
        & & Small & Medium \\
        \midrule
        \parbox[t]{2mm}{\multirow{4}{*}{\rotatebox[origin=c]{90}{\model{GPT2}}}}
        & Parameters & 209M & 355M \\
        & Hidden Size & 768 & 1024 \\
        & Intermediate Size & 3072 & 4096 \\
        & Hidden Layers & 24 & 24\\
        & Num Heads & 16 & 16 \\
        \midrule
        \parbox[t]{2mm}{\multirow{4}{*}{\rotatebox[origin=c]{90}{\model{xLSTM}}}}
        & Parameters & 247M & 406M \\
        & Hidden Size & 768 & 1024 \\
        & Intermediate Size & 2048 & 2731 \\
        & Hidden Layers & 24 & 24 \\
        & Num Heads & 4 & 4 \\
        \midrule
        \parbox[t]{2mm}{\multirow{4}{*}{\rotatebox[origin=c]{90}{\model{Mamba2}}}}
        & Parameters & 172M & 432M \\
        & Hidden Size & 768 & 1024 \\
        & Intermediate Size & 1536 & 2048 \\
        & Hidden Layers & 24 & 48 \\
        & Num Heads & 24 & 32 \\
        & State Size & 32 & 32 \\
        \midrule
        \parbox[t]{2mm}{\multirow{4}{*}{\rotatebox[origin=c]{90}{\model{LLaMA}}}}
        & Parameters & 208M & 360M \\
        & Hidden Size & 768 & 960 \\
        & Intermediate Size & 1536 & 2560 \\
        & Hidden Layers & 36 & 32 \\
        & Num Heads & 9 & 15 \\
        \bottomrule
    \end{tabularx}
    \vspace*{-2mm}
    \caption{Model configurations used during training.}\label{tab:model_params}
    \vspace*{-2mm}
\end{table}

\section{Further Results}
\label{sec:additional_results}

\begin{figure}[ht!]
\includegraphics[trim=0 5mm 0 8mm, width=\linewidth]{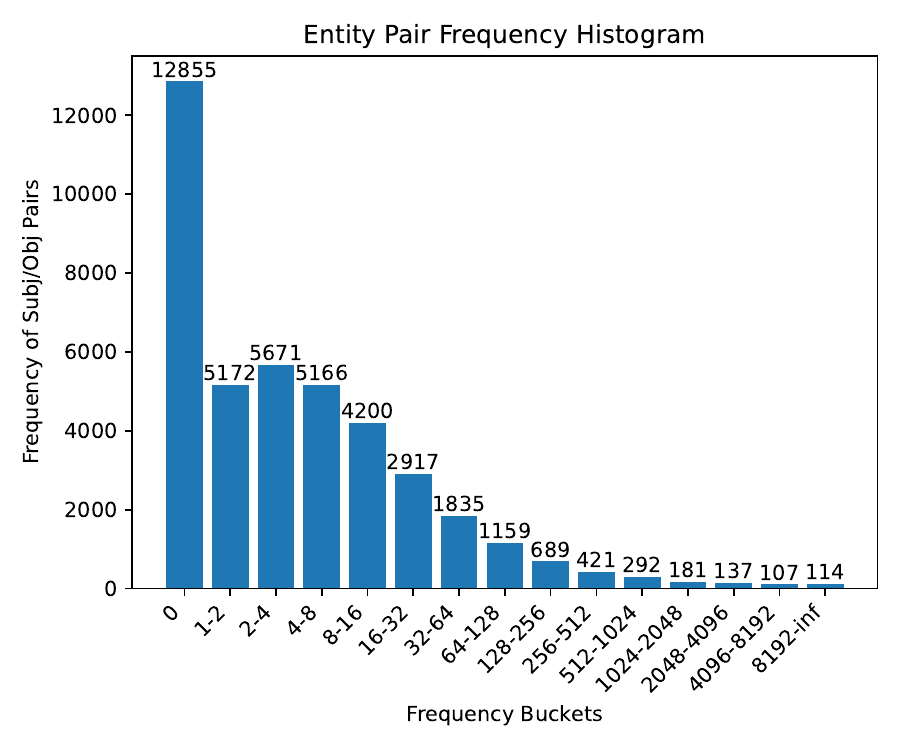}
\caption{Number of matches for BEAR-big facts in the English Wikipedia dump  \citep[\texttt{20231101.en}; ][]{wikidump}.}
\label{fig:approach_matching_results_bear_big}
\end{figure}

\begin{table}[ht!]
    \centering\small
\begin{tabularx}{\linewidth}{cX|cccc}
 \toprule
 & Model & \#params & ACC & WASB & $\alpha_m$\\
 \midrule
 \parbox[t]{2mm}{\multirow{4}{*}{\rotatebox[origin=c]{90}{\textsc{small}}}}
 &\model{GPT2}   & 209M & \underline{16.2\%} & 16.0\% & \underline{0.064}\\
 &\model{LLaMA}  & 208M & \textbf{18.2\%} &  \textbf{18.0\%} & \textbf{0.079}\\
 &\model{xLSTM}  & 247M & 15.6\% & 15.6\% & \underline{0.064}\\
 &\model{Mamba2} & 172M & 16.1\% & \underline{16.1\%} & \underline{0.064}\\
 \midrule
 \parbox[t]{2mm}{\multirow{4}{*}{\rotatebox[origin=c]{90}{\textsc{medium}}}}
 &\model{GPT2}   & 355M & 17.7\% & 17.5\% & 0.074\\
 &\model{LLaMA}  & 360M & \textbf{20.1\%} & \textbf{20.1\%} & \textbf{0.091}\\
 &\model{xLSTM}  & 406M & 17.3\% & 17.0\% & 0.073\\
 &\model{Mamba2} & 432M & \underline{18.5\%} & \underline{18.6\%} & \underline{0.080}\\
 \bottomrule
\end{tabularx}
\vspace*{-2mm}
    \caption{Results on BEAR-big.}
    \label{tab:results_bear_big}
\end{table}

\begin{table}[ht!]
    \centering\small
    \begin{tabularx}{\linewidth}{cX|ccc}
     \toprule
     &Model & \#params  &  $<$ 1024 & $\geq$ 1024\\
     \midrule
     \parbox[t]{2mm}{\multirow{4}{*}{\rotatebox[origin=c]{90}{\textsc{small}}}}
     &\model{GPT2}   & 209M & \underline{15.3\%} & \underline{79.9\%}\\
     &\model{LLaMA}  & 208M & \textbf{17.3\%} & \textbf{83.8\%}\\
     &\model{xLSTM}  & 247M  & 14.8\% & 77.9\%\\
     &\model{Mamba2} & 172M & \underline{15.3\%} & 77.3\%\\
     \midrule
     \parbox[t]{2mm}{\multirow{4}{*}{\rotatebox[origin=c]{90}{\textsc{medium}}}}
     &\model{GPT2}   & 355M & 16.8\% & \textbf{82.1\%}\\
     &\model{LLaMA}  & 360M & \textbf{19.3\%} & \underline{82.0\%}\\
     &\model{xLSTM}  & 406M  & 16.4\% & 79.5\%\\
     &\model{Mamba2} & 432M & \underline{17.7\%} & 79.4\%\\
     \bottomrule
    \end{tabularx}
    \vspace*{-2mm}
        \caption{Accuracy on high and low occurring facts on BEAR-big.}
        \label{tab:acc_high_low_facts_results_bear_big}
\end{table}

\begin{figure}[ht]
    \centering
    \begin{subfigure}{\linewidth}
        \includegraphics[trim=0 3mm 0 0, width=\linewidth]{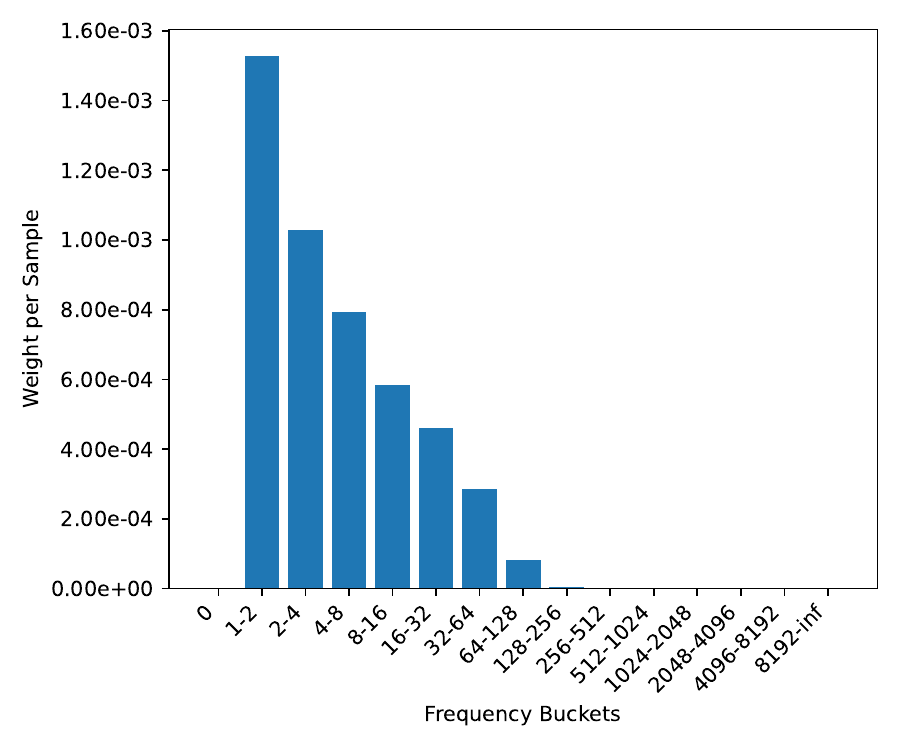}
        \caption{Weight impact for BEAR.}
        \label{fig:weight-impact-bear}
    \end{subfigure}
    \begin{subfigure}{\linewidth}
        \includegraphics[trim=0 3mm 0 0, width=\linewidth]{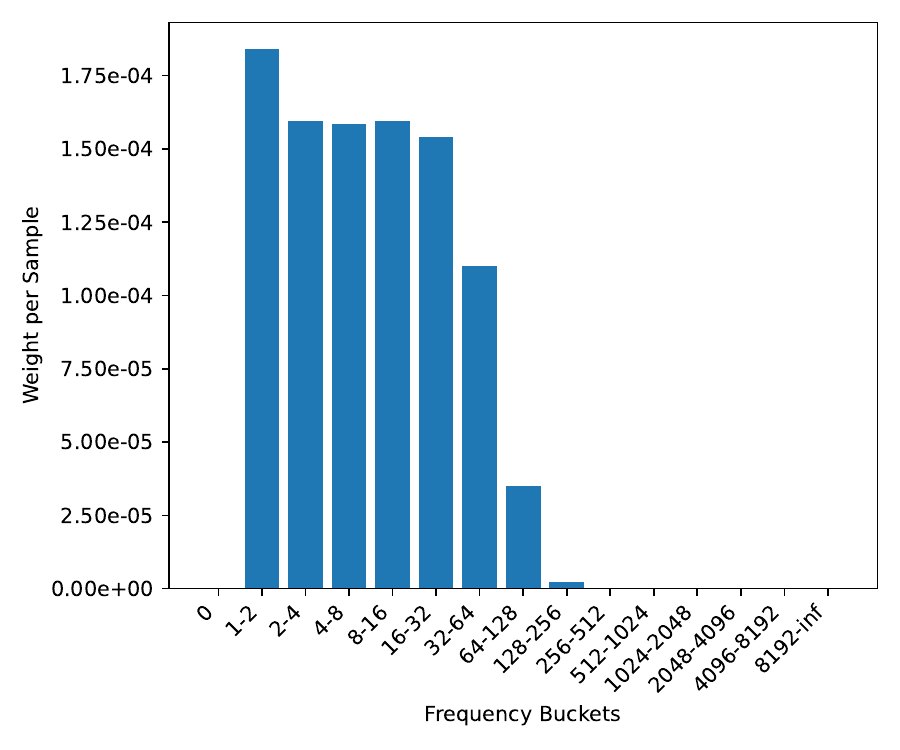}
        \caption{Weight impact for BEAR-big.}
        \label{fig:weight-impact-bear-big}
    \end{subfigure}
    \caption{Impact of the frequency bucket weight per number of samples.}
    \label{fig:weight-impact-bear-both}
\end{figure}

\begin{table*}[t]
        \centering\scriptsize
        \begin{tabularx}{\linewidth}{cX|cccccc}
         \toprule
         & Model & \#params & winogrande & wsc273 & lambada\_standard acc& lambada\_standard PPL & pile\_10k PPL\\
         \midrule
          \parbox[t]{2mm}{\multirow{4}{*}{\rotatebox[origin=c]{90}{\textsc{small}}}}
         &\model{GPT2} & 209M & 50.36\% $\pm$ 1.4\% & 53.11\% $\pm$ 3.03\% & 16.63\% $\pm$ 0.52\% & 652.0058 $\pm$ 33.1575 & 14389.4299\\
         &\model{LLaMA} & 208M & 50.59\% $\pm$ 1.4\% & 55.68\% $\pm$ 3.01\% & 15.58\% $\pm$ 0.51\% & 694.1146 $\pm$ 34.3843 & 65059.5665\\
         &\model{xLSTM} & 247M & 50.43\% $\pm$ 1.4\% & 54.95\% $\pm$ 3.02\% & 9.35\% $\pm$ 0.41\% & 1536.1172 $\pm$ 74.8833 & 966.7574\\         
         &\model{Mamba2} & 172M & 50.2\% $\pm$ 1.4\% & 50.92\% $\pm$ 3.03\% & 7.68\% $\pm$ 0.37\% & 2183.7652 $\pm$ 109.3855 & 1295.2241\\
          \midrule
         \parbox[t]{2mm}{\multirow{4}{*}{\rotatebox[origin=c]{90}{\textsc{medium}}}}
         &\model{GPT2} & 355M & 51.62\% $\pm$ 1.4\% & 54.58\% $\pm$ 3.02\% & 16.44\% $\pm$ 0.52\% & 592.8151 $\pm$ 29.6474 & 17984.4641\\
         &\model{LLaMA} & 360M & 51.85\% $\pm$ 1.4\% & 54.58\% $\pm$ 3.02\% & 15.76\% $\pm$ 0.51\% & 508.1769 $\pm$ 23.8731 & 216732.2782\\
         &\model{xLSTM} & 406M & 51.46\% $\pm$ 1.4\% & 50.55\% $\pm$ 3.03\% & 11.97\% $\pm$ 0.45\% & 739.1623 $\pm$ 34.8244 & 890.4901\\
         &\model{Mamba2} & 432M & 50.67\% $\pm$ 1.4\% & 54.58\% $\pm$ 3.02\% & 7.88\% $\pm$ 0.38\% & 1594.1999 $\pm$ 77.5151 & 1116.7870\\
         \bottomrule
        \end{tabularx}
            \caption{LM Evaluation Harness Results.}
            \label{tab:lm_eval_harness_results}
\end{table*}

\begin{figure*}[ht]
\includegraphics[width=\linewidth]{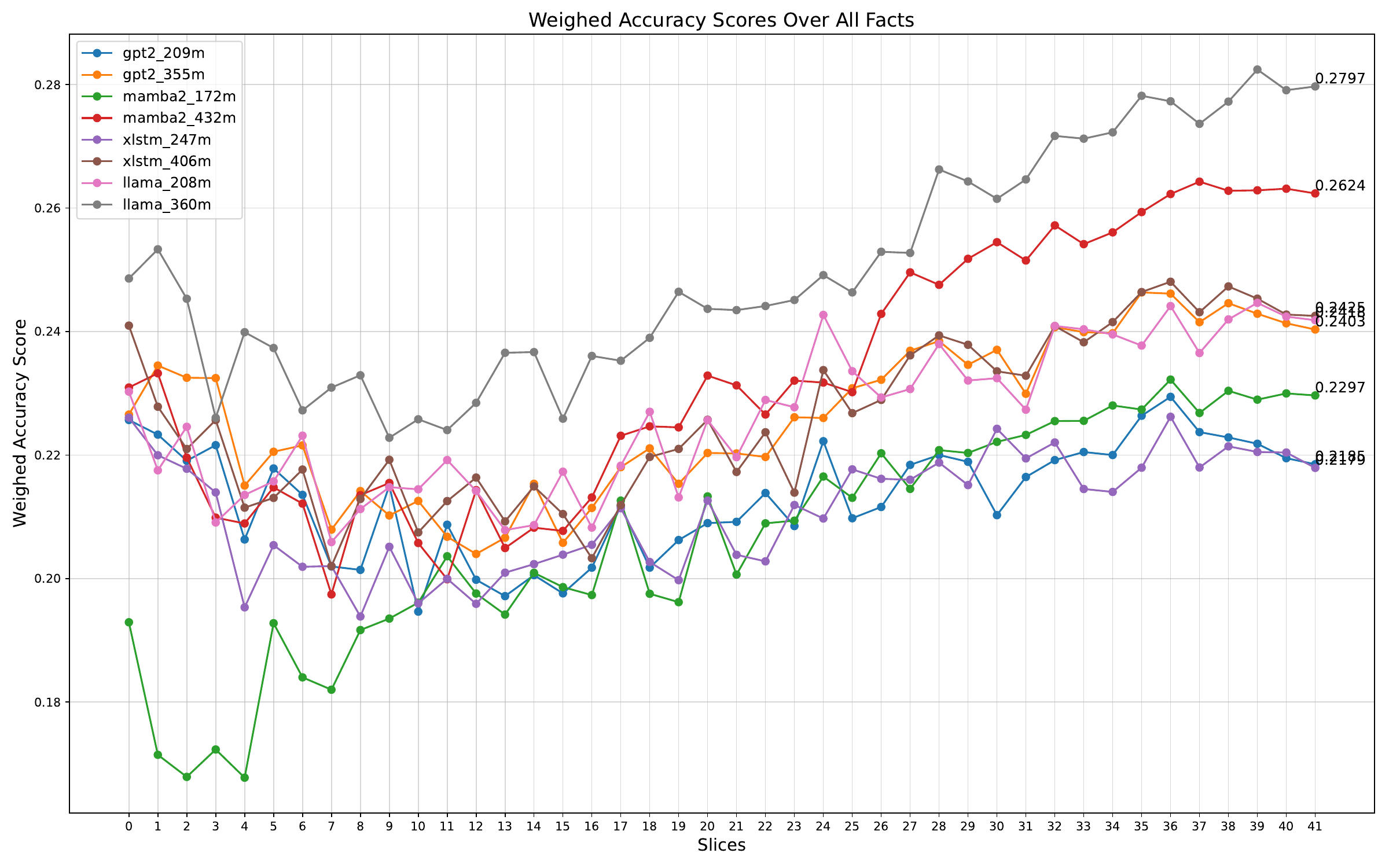}
\caption{Development of the weighted accuracy (WASB) throughout the pre-training.}
\label{fig:weighted_accuracy_on_slices_bear}
\end{figure*}

\begin{figure*}[ht]
\includegraphics[width=\linewidth]{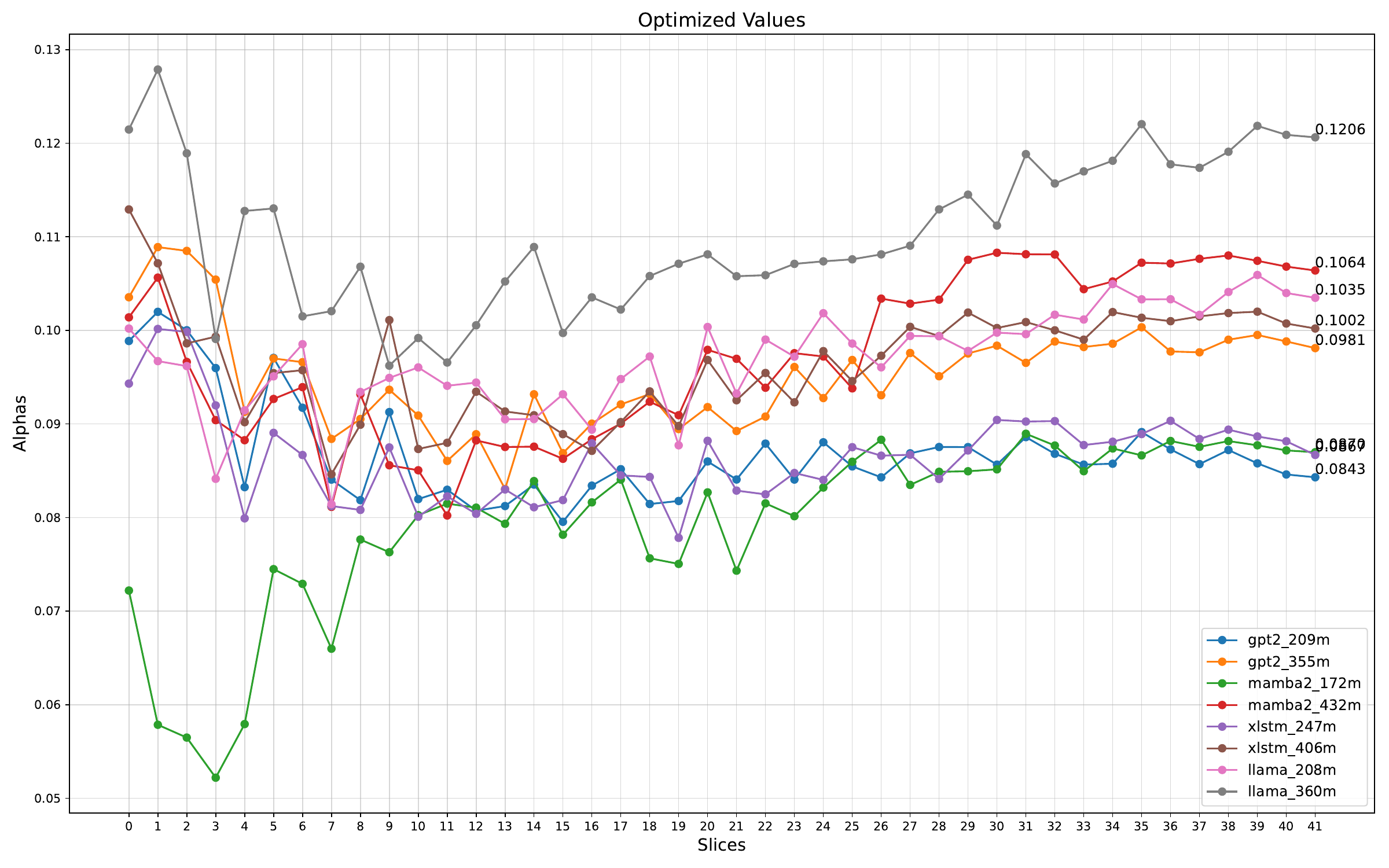}
\caption{Development of $\alpha_m$ over the course of the pre-training.}
\label{fig:psf-ext3_optimized_alphas_bear}
\end{figure*}

\begin{figure*}[ht]
\includegraphics[width=\linewidth]{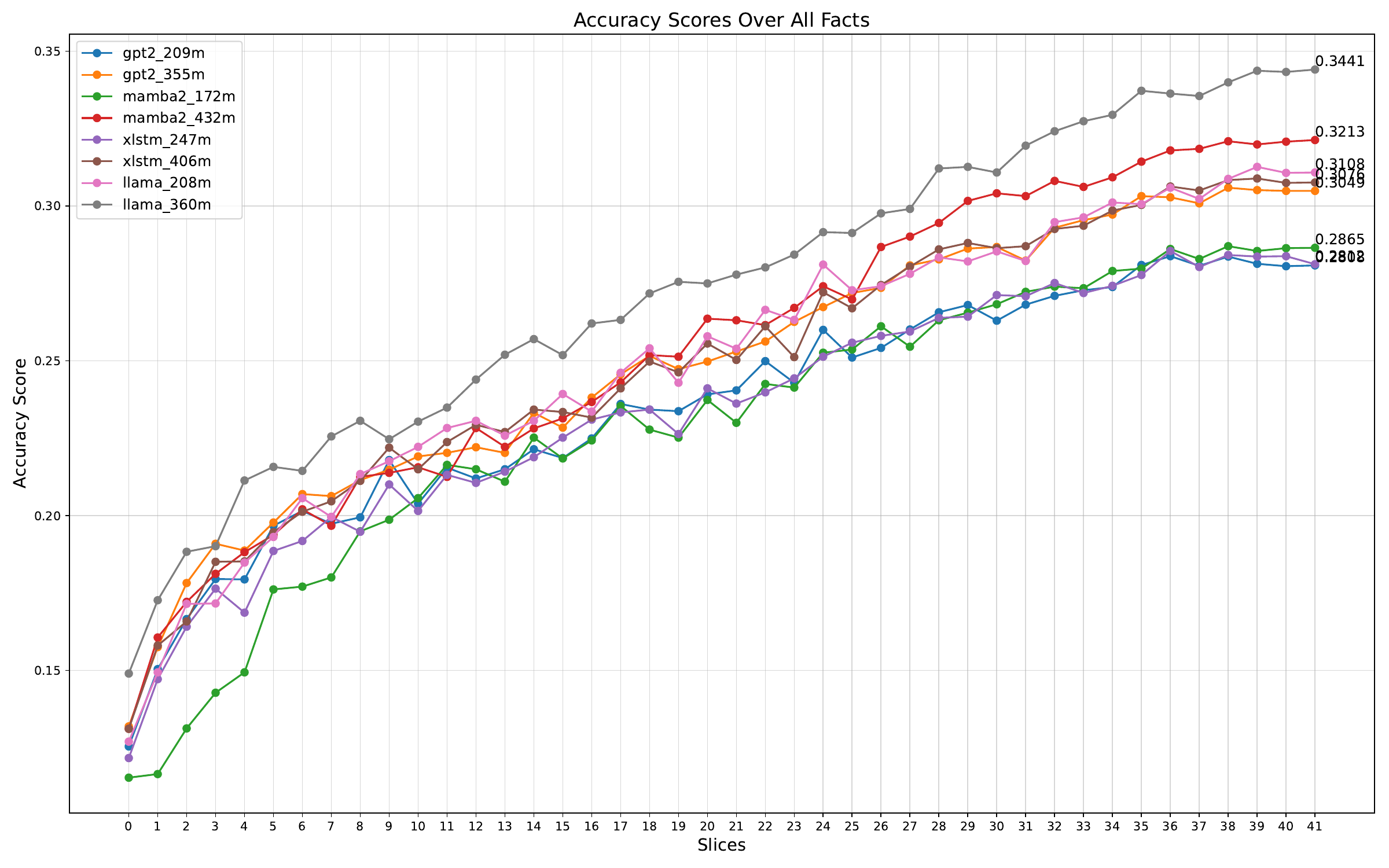}
\caption{Development of the accuracy throughout the pre-training. }
\label{fig:accuracy_on_slices_bear_small}
\end{figure*}

\begin{figure*}[ht]
\includegraphics[width=\linewidth]{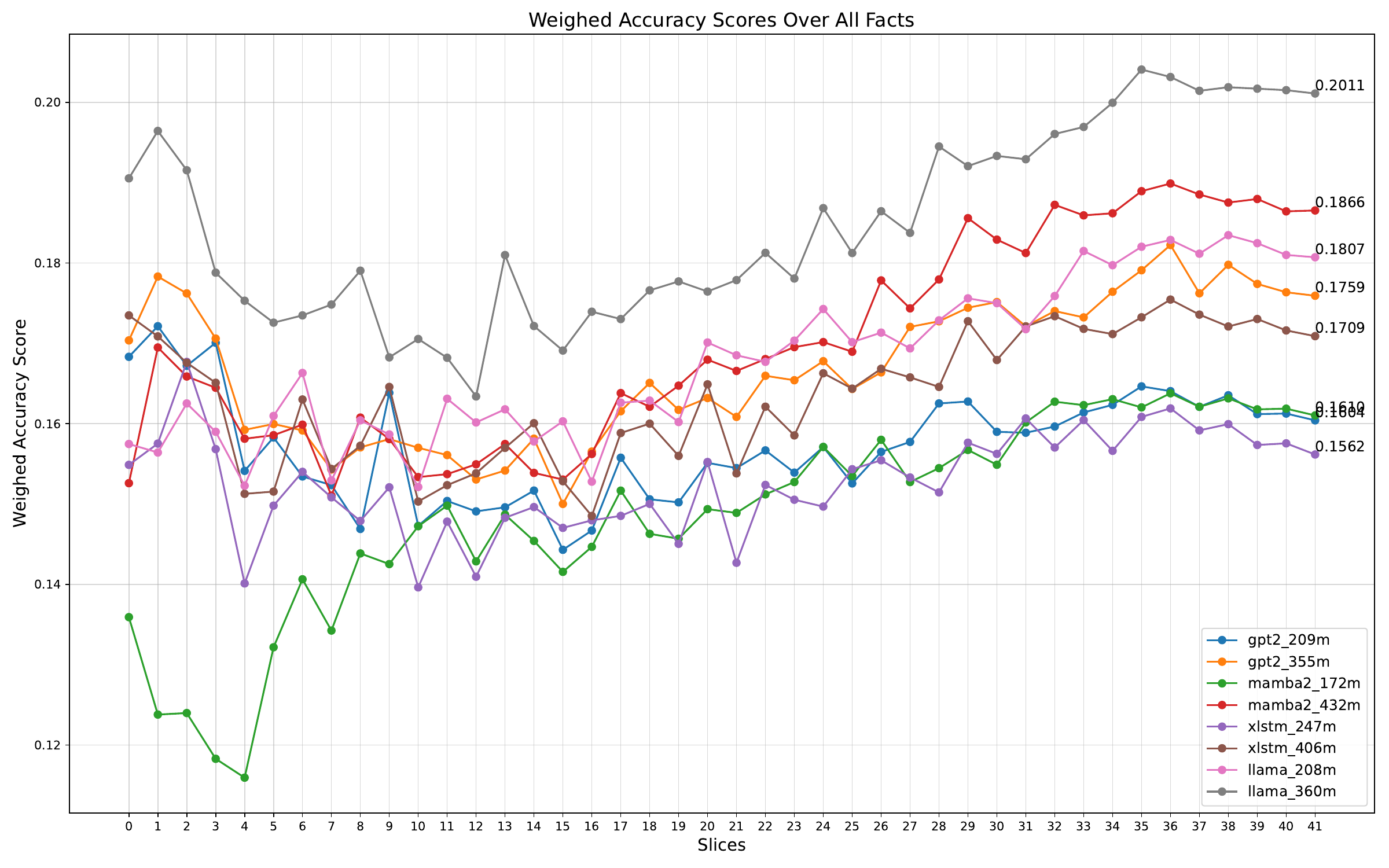}
\caption{Development of the weighted accuracy (WASB) throughout the pre-training as measured on BEAR-big.}
\label{fig:weighted_accuracy_on_slices_bear_big}
\end{figure*}

\begin{figure*}[ht]
\includegraphics[width=\linewidth]{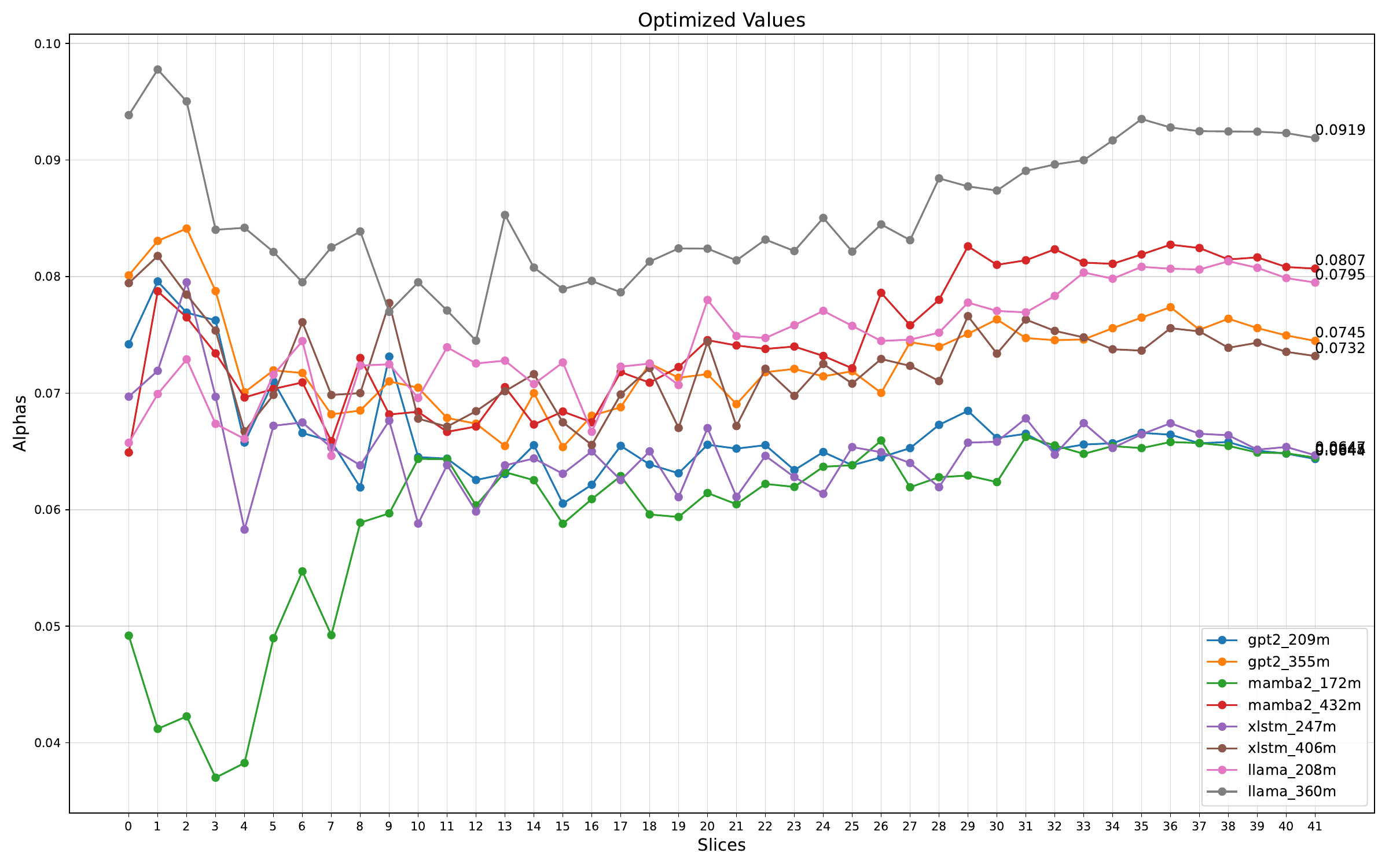}
\caption{Development of $\alpha_m$ throughout the pre-training as measured on BEAR-big.}
\label{fig:psf-ext3_optimized_alphas_bear_big}
\end{figure*}

\begin{figure*}[ht]
\includegraphics[width=\linewidth]{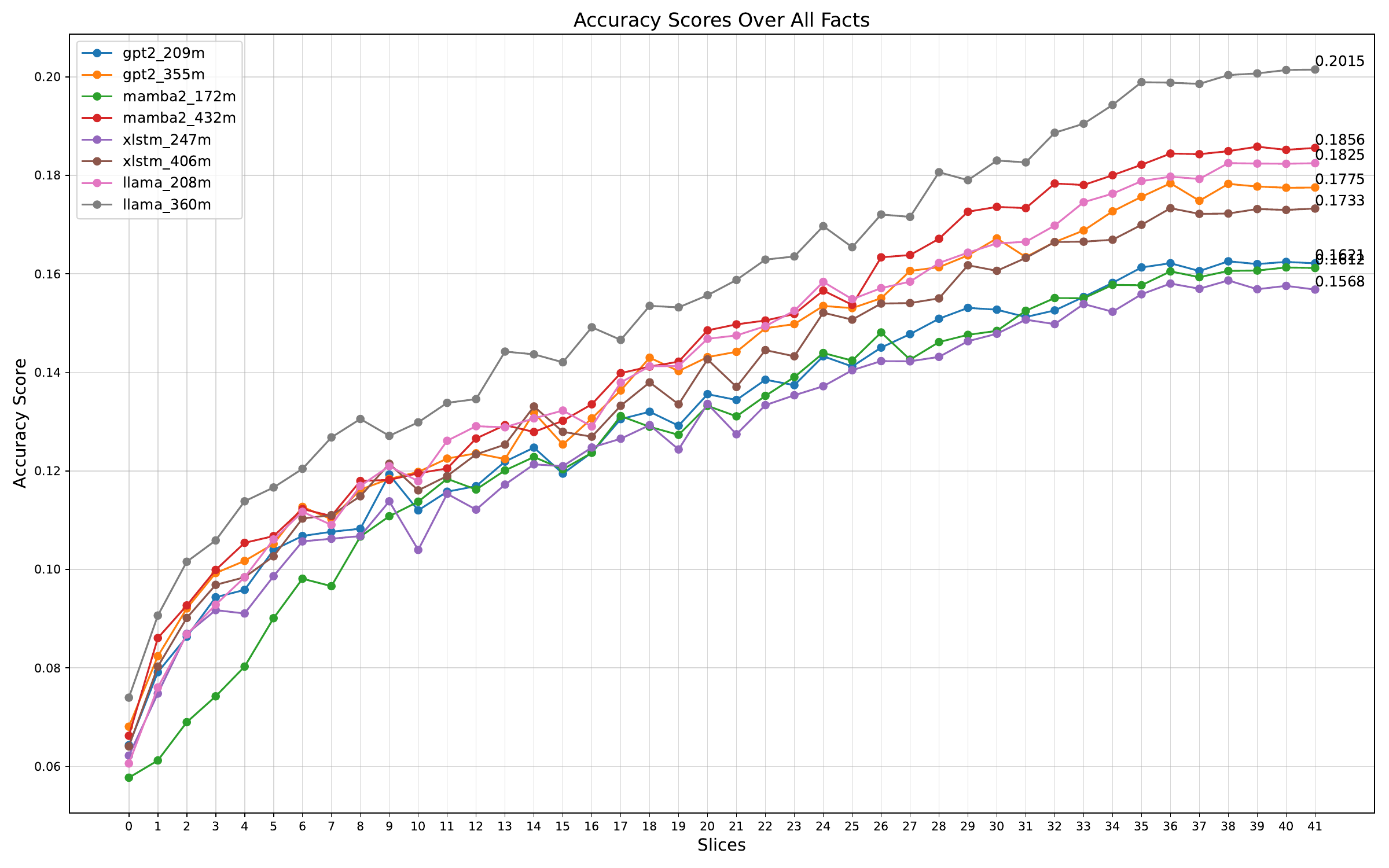}
\caption{Development of the accuracy throughout the pre-training as measured on BEAR-big. }
\label{fig:accuracy_on_slices_bear_big}
\end{figure*}

\begin{figure*}[t!]
    \centering
    \begin{subfigure}{0.49\linewidth}
        \includegraphics[width=\linewidth]{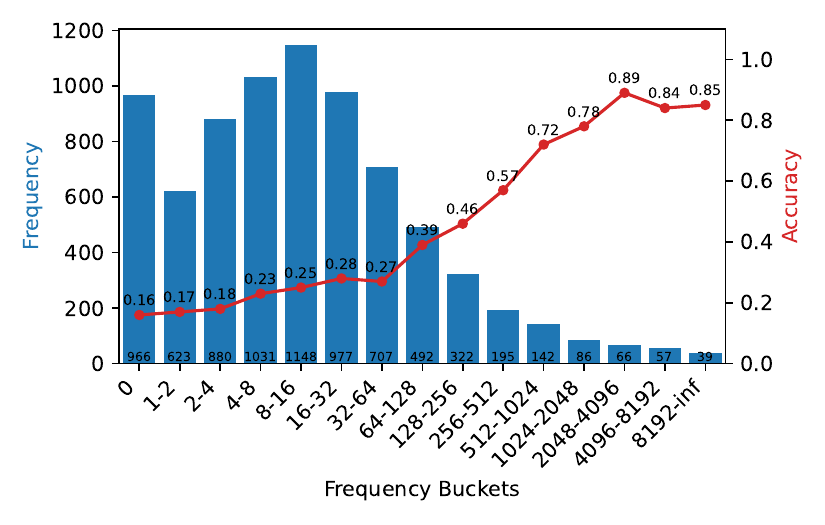}
        \caption{\model{GPT2} 209m.}
    \end{subfigure}
    \hfill
    \begin{subfigure}{0.49\linewidth}
        \includegraphics[width=\linewidth]{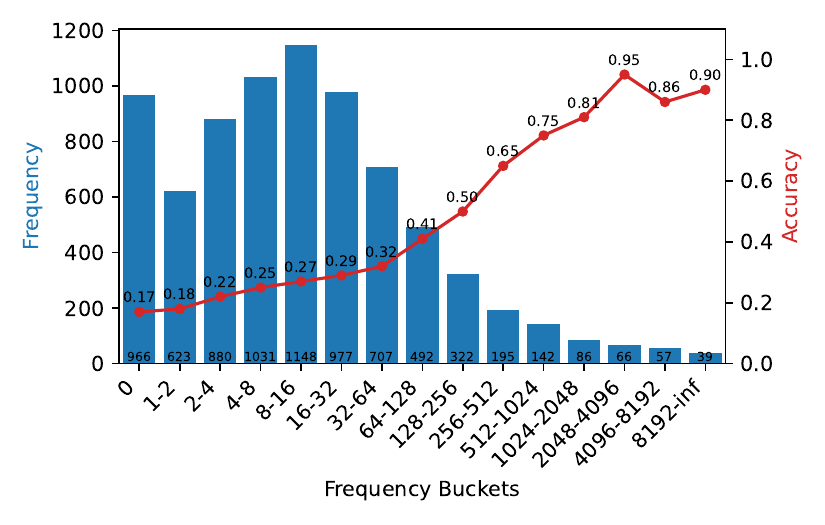}
        \caption{\model{GPT2} 355m.}
    \end{subfigure}
    \hfill
    \vspace{0.5cm}
    \hfill
    \begin{subfigure}{0.49\linewidth}
        \includegraphics[width=\linewidth]{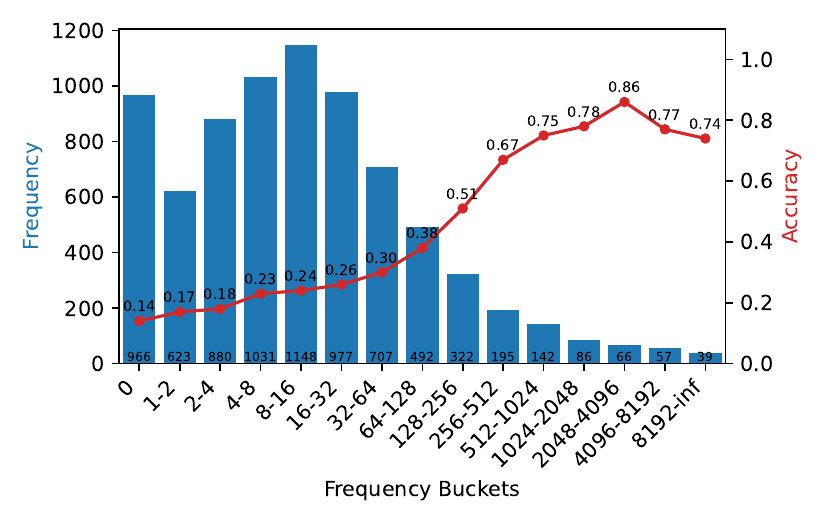}
        \caption{\model{xLSTM} 247m.}
    \end{subfigure}
    \hfill
    \begin{subfigure}{0.49\linewidth}
        \includegraphics[width=\linewidth]{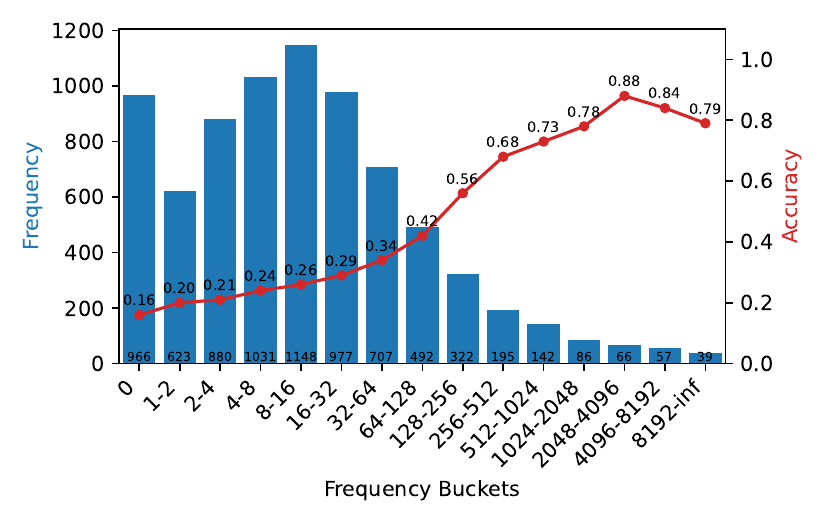}
        \caption{\model{xLSTM} 406m.}
    \end{subfigure}
    \hfill
    \vspace{0.5cm} 
    \hfill
    \begin{subfigure}{0.49\linewidth}
        \includegraphics[width=\linewidth]{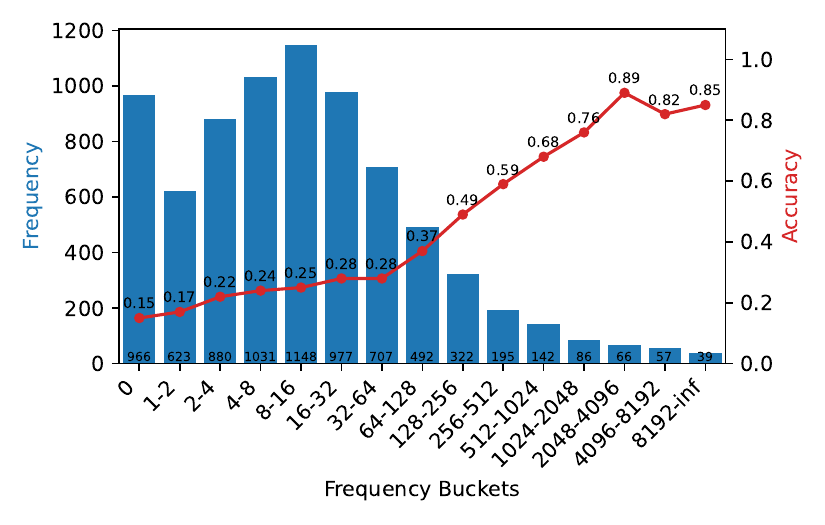}
        \caption{\model{Mamba2} 172m.}
    \end{subfigure}
    \hfill
    \begin{subfigure}{0.49\linewidth}
        \includegraphics[width=\linewidth]{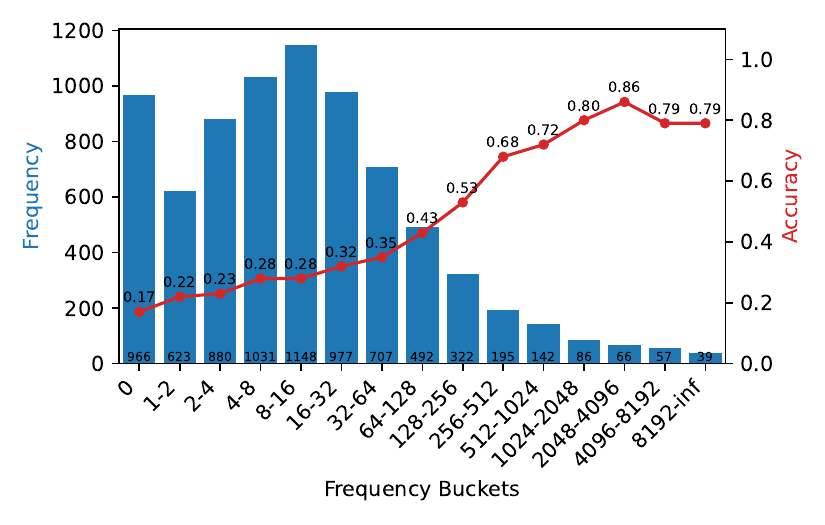}
        \caption{\model{Mamba2} 432m.}
    \end{subfigure}
    \hfill
    \vspace{0.5cm} 
    \hfill
    \begin{subfigure}{0.49\linewidth}
        \includegraphics[width=\linewidth]{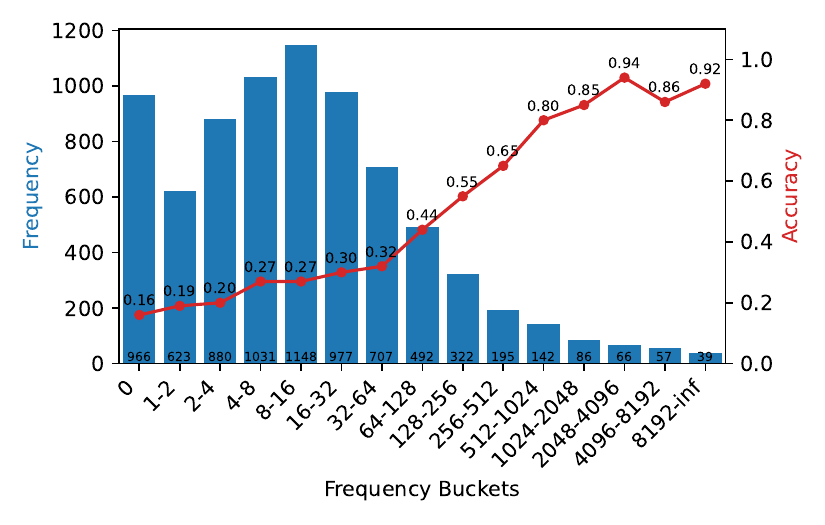}
        \caption{\model{LLaMA} 208m.}
    \end{subfigure}
    \hfill
    \begin{subfigure}{0.49\linewidth}
        \includegraphics[width=\linewidth]{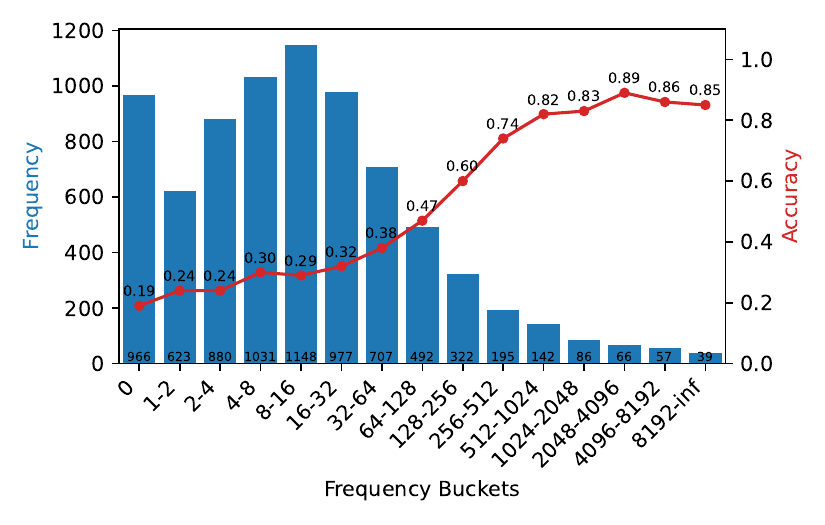}
        \caption{\model{LLaMA} 360m.}
\end{subfigure}

    \caption{Frequency Bucket Accuracy of the model's final state as measured on BEAR.}
    \label{fig:final_bucket_acc_all_models}
\end{figure*}

\begin{figure*}[t]
    \centering
    \begin{subfigure}{\textwidth}
        \centering
        \includegraphics[width=0.8\linewidth]{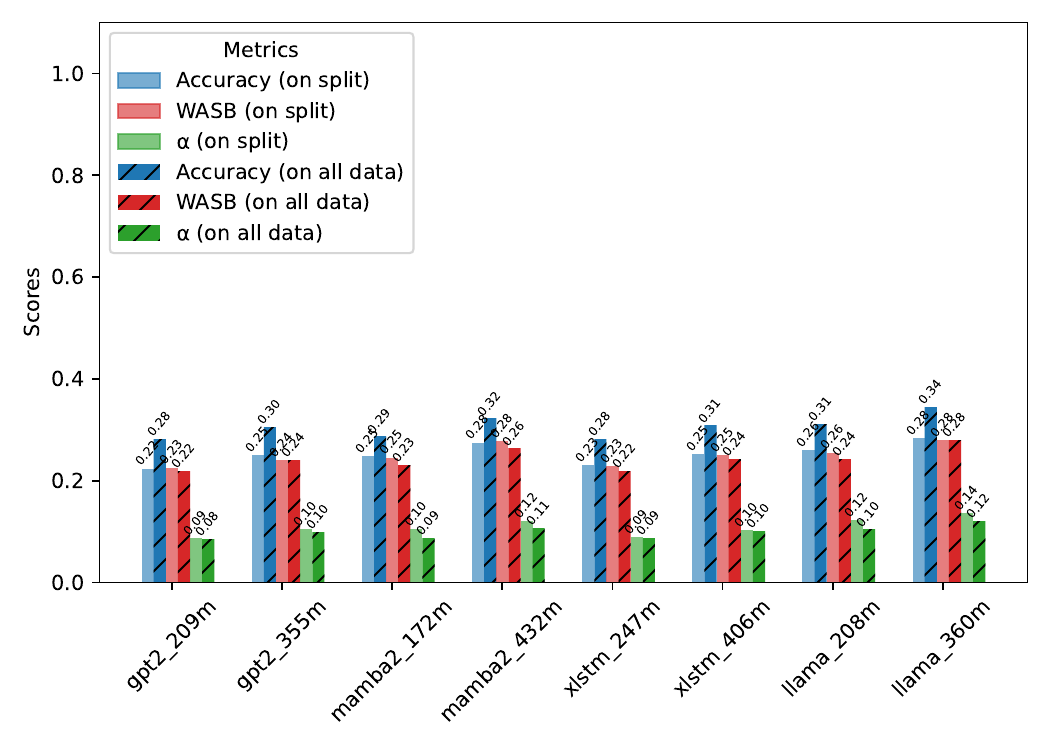}
        \caption{Low frequency-split}
        \label{fig:all-metrics--for-low-frequency-split}
    \end{subfigure}
    \begin{subfigure}{\textwidth}
        \centering
        \includegraphics[width=0.8\linewidth]{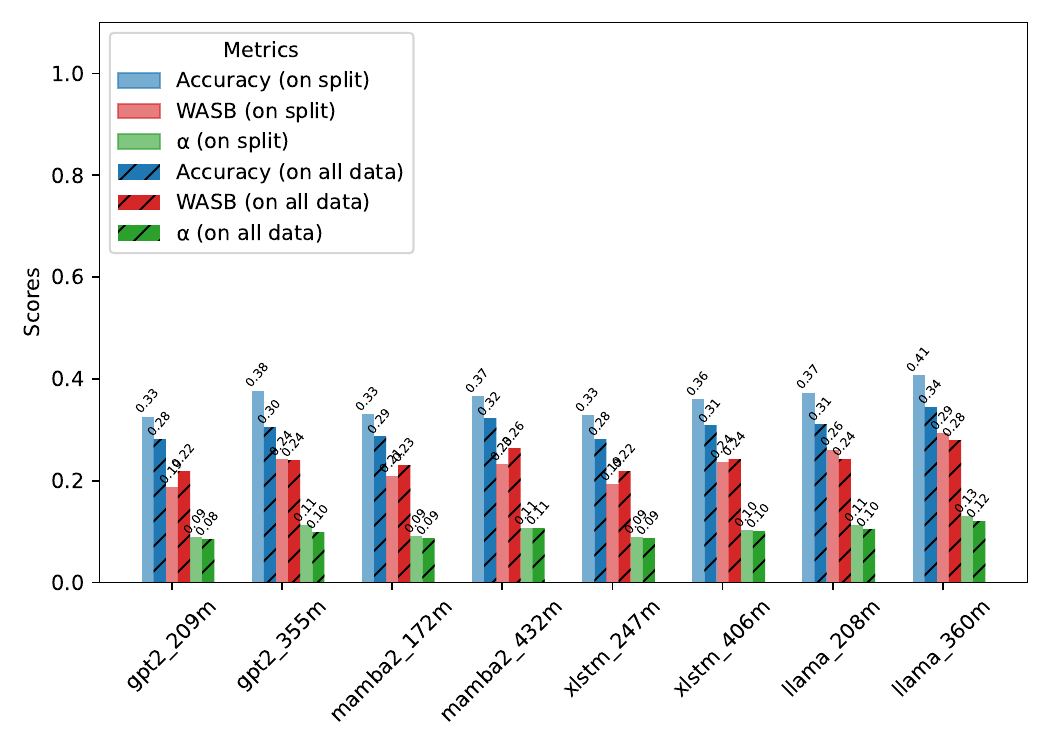}
        \caption{High frequency-split}
        \label{fig:all-metrics-for-high-frequency-split}
    \end{subfigure}
    \caption{Accuracy, WASB and $\alpha$ scores on the low and high frequency splits and entire data set for comparison on BEAR.}
    \label{fig:all-metrics-for-frequency-splits}
\end{figure*}

\end{document}